\pdfoutput=1
\documentclass[10pt,twocolumn,letterpaper]{article}
\PassOptionsToPackage{table}{xcolor}

\usepackage[pagenumbers]{cvpr}

\usepackage[T1]{fontenc}
\usepackage{microtype}
\usepackage{colortbl}
\usepackage{algorithm}
\usepackage[noend]{algpseudocode}
\algrenewcommand{\algorithmiccomment}[1]{\hfill\textcolor{gray}{$\triangleright$\,#1}}\algnewcommand{\LineComment}[1]{\State\textcolor{gray}{$\triangleright$\,#1}}

\definecolor{cvprblue}{rgb}{0.21,0.49,0.74}
\usepackage[pagebackref,breaklinks,colorlinks,allcolors=cvprblue]{hyperref}
\usepackage{bm}
\colorlet{ivtbase}{blue!6}   
\colorlet{ivtstride}{blue!15} 

\def\confName{CVPR}
\def\confYear{2027}

\newcommand{\STRIDE}{STRIDE}
\newcommand{\apivt}{AP$_\mathrm{IVT}$}

\renewcommand{\paragraph}[1]{\vspace{2pt}\noindent\textbf{#1}\hspace{0.5em}}

\title{A Closed-Form Temporal Readout for Surgical Triplet Recognition}
\author{Lemar Abdi$^{1}$\thanks{Corresponding author: \texttt{l.abdi@tue.nl}} \quad Amaan Valiuddin$^{1}$ \quad Fida Mohammad Thoker$^{2}$\\
Karen Sanchez$^{2}$ \quad Bernard Ghanem$^{2}$ \quad Fons van der Sommen$^{1}$\\
$^{1}$Eindhoven University of Technology \qquad $^{2}$King Abdullah University of Science and Technology\vspace{-14pt}}

\definecolor{dup}{HTML}{2E8B57}
\definecolor{stride}{HTML}{EAF2FB}



\begin{document}
\maketitle
\begin{abstract}
Surgical triplet recognition requires identifying which instrument performs which action on which anatomical target, and following these interactions as they unfold over time. The task is challenging because several triplets can occur at once, their components span very different durations, and the evidence for an interaction is rarely complete in a single frame. We introduce STRIDE, a post-hoc temporal readout that adds temporal context to any trained recogniser. It averages the recogniser's features and logits over a few fixed temporal windows and combines them with the current frame through ridge regression, solved in closed form on a CPU. Applied as a plug-and-play readout to nine existing methods, STRIDE improves \apivt{} in 24 of 27 method-dataset pairs, by up to 7.4 points. Moreover, we propose a simple pipeline consisting of an adapted foundation vision encoder with a linear classifier and STRIDE. Across thirteen encoders, STRIDE improves 37 of 39 encoder-dataset pairs by up to 7.2 points. %
STRIDE reduces false alarms by 41\%, at the cost of a slower response at interaction boundaries.
Code will be made publicly available upon acceptance.
\end{abstract}
\section{Introduction}
\label{sec:intro}

Computer-assisted surgery requires an understanding of fine-grained activity in the surgical scene. Surgical triplet recognition aims to identify, at every frame, which \emph{instrument} performs which \emph{verb} on what \emph{target}~\cite{nwoye2020tripnet}. A single frame often cannot answer this. An instrument can keep dissecting tissue while hiding the point of contact, and verbs such as grasping and retracting can look identical in a still image. Earlier frames may show the target more clearly or reveal the motion that tells such actions apart~\cite{sharma2023rit}, so recognition depends on temporal context. How much context is useful varies, because brief actions occur alongside longer interactions. During surgery, this choice is harder, as only past frames are available for online recognition~\cite{li2024survey}.

Existing methods use specialised architectures to model spatial cues, temporal dependencies, and relations among triplet components~\cite{sharma2023rit,sharma2026spirit,li2026trcorsurg,jeon2026curconmixplus}. They differ in how they combine evidence, yet all of them produce frame-level visual features and class logits. This shared interface raises a question: \textit{can we use the features and logits that a recogniser already produces to improve its predictions after training?}

To answer this, we introduce STRIDE, a post-hoc temporal readout for surgical triplet recognition. At each frame, STRIDE combines the recogniser's current features and logits with their averages over several temporal windows. A linear readout maps this representation to triplet and component scores. We fit the readout stage with closed-form ridge regression in minutes on a CPU, while keeping the recogniser fixed. The same temporal scales are used across recognisers, with causal windows for online recognition and centred windows for offline recognition.
Furthermore, this design allows us to assess how well a simple recogniser can perform on this task. We jointly adapt an image or video foundation encoder with a linear classifier, then fit STRIDE on their features and logits using the same procedure.

STRIDE improves triplet recognition in 24 of the 27 recogniser--dataset pairs, with the largest gains for framewise models. Combined with a foundation encoder and a linear classifier, STRIDE provides a competitive online baseline. Offline, this combination exceeds the specialised recognisers in our comparison, with an \apivt{} of 45.1 on CholecT45, 43.0 on CholecT50 and 44.6 on ProstaTD. Qualitative examples show that STRIDE can maintain triplet predictions during brief occlusions and reduce false positives, although predictions can persist after an interaction ends. These findings show that a post-hoc readout can recover useful temporal context and that, paired with STRIDE and a strong visual encoder, even a linear classifier achieves competitive triplet recognition performance. Our contributions are as follows:
\begin{itemize}
\item We introduce STRIDE, a closed-form post-hoc readout that adds multi-scale temporal context to trained surgical triplet recognisers in both online and offline settings.

\item We evaluate it across nine existing recognisers and thirteen foundation encoders on three datasets, and use ablations and segment-level analysis to examine where the gains come from and how they affect the timing of predictions.
\item We show that an adapted foundation vision encoder with a linear classifier and STRIDE is competitive with or surpasses specialised triplet recognition methods.
\end{itemize}

\section{Related work}
\label{sec:related}

\paragraph{Surgical action triplet recognition.}
Surgical action triplets describe each tool--tissue interaction as an instrument, a verb, and a target. Tripnet~\cite{nwoye2020tripnet} first recognised them by guiding the verb and target predictions with instrument activations.  Rendezvous~\cite{nwoye2022rendezvous} extended this with attention over instrument activation maps. Subsequent methods localise interactions with text-conditioned or multi-scale spatial features~\cite{sharma2026spirit,li2026trcorsurg}. Others add temporal context, through clip-level attention~\cite{sharma2023rit}, multi-stage convolutions~\cite{gui2024terl}, a causal transformer denoiser~\cite{liu2024difftriplet}, multi-resolution transformers~\cite{jeon2026curconmixplus} or frame memories~\cite{li2026trcorsurg}. The triplet components are further related by disentangling them~\cite{chen2023tripdis}, by graphs~\cite{xi2022forestgcn,li2026trcorsurg,sharma2026spirit} or by priors from language models~\cite{xi2023chainoflook,mejo2025,moeco2026,surgmllm2026}. Training is further improved through distillation~\cite{yamlahi2023sdswin,gui2024mt4mtlkd} and contrastive learning~\cite{gui2024terl,jeon2025curconmix}. Despite their differences, these approaches typically introduce task-specific components to model spatial, temporal or relational structure.

\paragraph{Temporal modelling in untrimmed videos.}
Temporal models are commonly trained on per-frame features, using multi-stage convolutions~\cite{farha2019mstcn,li2020mstcnpp}, transformers~\cite{yi2021asformer}, state-space models~\cite{sinha2025mstemba} or diffusion~\cite{liu2023diffact}. Surgical phase recognition followed the same path, using HMMs and LSTMs on frame features~\cite{twinanda2017endonet,jin2021tmrnet}, temporal convolutions~\cite{czempiel2020tecno} and transformers~\cite{gao2021transsvnet,liu2025lovit,yang2024surgformer}, often trained in a separate stage on cached features~\cite{rivoir2024pitfalls}. In these methods, the temporal module is learned for a specific architecture. In general video understanding, some approaches aggregate stored features or predictions around the current frame. Long-term feature banks~\cite{wu2019lfb} and online action detectors~\cite{xu2021lstr,zhao2022testra} aggregate past features over several horizons. Refinement networks correct per-frame predictions~\cite{ishikawa2021asrf,ahn2021hasr}, and stacked learning feeds the predictions of a fixed model to a second learner~\cite{wolpert1992stacked,cohen2005stacked}. These approaches train an additional model for the aggregation. Linear readouts can be fitted in closed form by ridge regression~\cite{alain2016probes,dempster2020rocket,mcdonnell2023ranpac}, but are typically applied to the features of individual samples. STRIDE instead applies to any trained recogniser, is fitted in closed form without training an additional model, and reads both the features and the predictions of the recogniser over several temporal windows.

\paragraph{Adapting vision foundation models to surgical data.}
General-purpose image encoders~\cite{simeoni2025dinov3,tschannen2025siglip2} and video encoders~\cite{wang2023videomaev2,wang2024internvideo2,assran2025vjepa2,vandeghen2026trackmae} are commonly probed with frozen features or adapted with parameter-efficient fine-tuning~\cite{hu2022lora,yuan2024videoglue}. For surgical data, such encoders are either adapted directly with parameter-efficient modules~\cite{yang2026surgpetl} or pre-trained on surgical images and videos. This pre-training is self-supervised, through masked image modelling~\cite{batic2024endovit}, self-distillation~\cite{jaspers2025surgenetxl,che2026lemon} or temporal-order ranking~\cite{che2026plstitch}, or uses video--language supervision~\cite{yuan2024peskavlp,che2026surglime}. Surgical video encoders have recently been scaled up as well~\cite{lu2026surgrec,unisurg2026}. Downstream, these encoders are typically evaluated with a separately trained temporal model on top of their frame features~\cite{jaspers2025surgenetxl,che2026lemon}, and with little data it matters where the temporal context is placed~\cite{strater2026temporal}. We instead adapt image and video encoders with a single linear classifier. Temporal context is added only through STRIDE.

\begin{figure*}[t]
    \centering
    \includegraphics[width=0.8
    \textwidth]{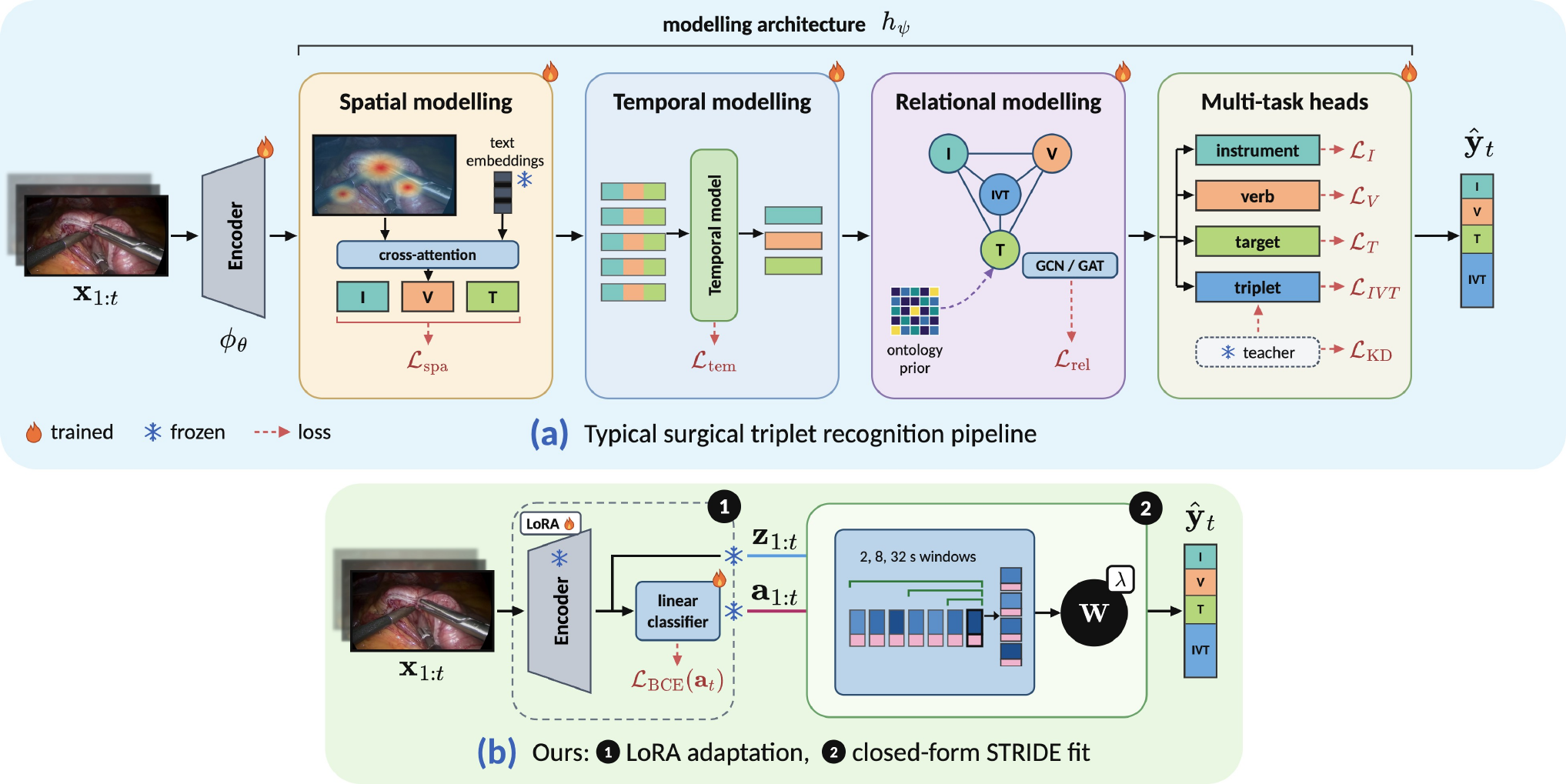}
\caption{\textbf{(a)} A typical surgical triplet recognition pipeline and \textbf{(b)} our foundation-model pipeline with STRIDE.}
    \label{fig:X}
\end{figure*}

\section{Method}

\subsection{Preliminaries}
\label{sec:typical}
Given a surgical video $X=\{\mathbf{x}_t\}_{t=1}^{L}$, surgical triplet recognition identifies, at each frame $t$, which instrument (I) performs which verb (V) on what target (T). Let $\mathcal{C}$ denote the vocabulary of $C$ triplets, where each triplet ${c}$ comprises an instrument $i({c})$, a verb $v({c})$, and a target $\tau({c})$. The model aims to predict a score vector $\hat{\mathbf{y}}^\mathrm{IVT}_t\in\mathbb{R}^{C}$ over these triplets. Multiple triplets may occur simultaneously, making this a multi-label recognition task. In the online setting, $\hat{\mathbf{y}}^\mathrm{IVT}_t$ must be predicted using only the frames $\mathbf{x}_{1:t}$ observed up to time $t$, whereas the offline setting may use the full video $\mathbf{x}_{1:L}$.

A general framework for surgical triplet recognition consists of a pre-trained visual encoder $\phi_\theta$ and a specialised head $h_\psi$, as shown in \cref{fig:X}a.  
The head uses the encoder's feature maps to model one or more of the following: spatial cues using instrument-guided activation maps~\cite{nwoye2022rendezvous}, temporal context via temporal attention or multi-resolution transformers~\cite{sharma2023rit,jeon2026curconmixplus}, and dependencies among instruments, verbs, and targets~\cite{nwoye2020tripnet,li2026trcorsurg,sharma2026spirit}. Classification layers then produce triplet scores and individual component scores. These architectures differ in how they construct and combine recognition evidence, yet their trained models all produce frame-level visual representations and class scores.
In the online setting, the triplet scores are given by
\begin{equation}
    \hat{\mathbf{y}}^\mathrm{IVT}_t = h_\psi\big(\phi_\theta(\mathbf{x}_{1:t})\big).
    \label{eq:typical}
\end{equation}
However, an ongoing interaction can be difficult to recognise from a single frame when the instrument is partially occluded or its action is ambiguous~\cite{sharma2023rit}. In such cases, representations and scores from earlier frames can provide context.

\subsection{STRIDE: a post-hoc temporal readout}
\label{sec:stride}

\begin{figure}[htbp]
    \centering \includegraphics[width=\linewidth]{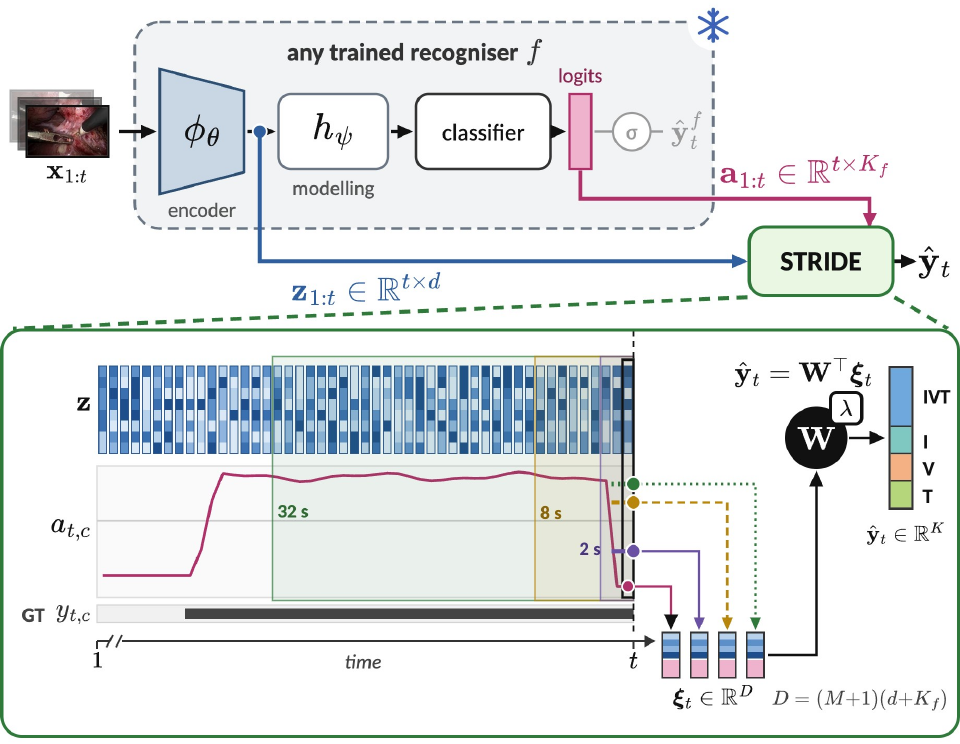}
    \caption{\textbf{STRIDE} on a fixed recogniser $f$. Features $\mathbf{z}_t$ and logits $\mathbf{a}_t$ are concatenated with their means over $M$ temporal windows to form $\bm{\xi}_t$. The ridge readout $\mathbf{W}$ maps $\bm{\xi}_t$ to triplet and component scores $\hat{\mathbf{y}}_t$. The lower panel shows one feature column and the logit $a_{t,c}$ and label $y_{t,c}$ of one class $c$.}
    \label{fig:Y}
\end{figure}

Surgical interactions unfold over time, but their visual evidence is not equally reliable at every frame. 
Earlier features and predictions can provide supporting evidence for recognising the current interaction. The useful extent of this history can also vary, since short windows retain recent evidence, whereas longer windows provide broader context~\cite{jeon2026curconmixplus}. Prior work shows that recognition outputs can serve as contextual inputs to a subsequent predictor~\cite{cohen2005stacked,ahn2021hasr}. Following this principle, we introduce STRIDE, a post-hoc temporal readout for surgical triplet recognition. At each frame, we concatenate the features and logits of a trained recognition model with their means over several temporal windows. We fit a linear readout on this representation with a ridge regression to predict triplet and component scores. The readout is fitted in closed form while the recognition model remains fixed. \Cref{fig:Y} shows an overview of STRIDE.

\paragraph{Step 1: per-frame evidence.} Given any trained recogniser $f$ with output dimensionality $K_f$, we extract a pooled visual representation $\mathbf{z}_t\in\mathbb{R}^{d}$ and classification logits $\mathbf{a}_t\in\mathbb{R}^{K_f}$. We then concatenate them as $\mathbf{u}_t = [\mathbf{z}_t; \mathbf{a}_t]$.

\paragraph{Step 2: window means.} We average $\mathbf{u}_t$ over $M$ temporal windows of radius $\omega_j$. As all videos are sampled at 1\,fps, each $\omega_j$ is given in seconds and equals the corresponding number of frames. For online recognisers the window $\mathcal{I}_t^{(j)}$ holds frame $t$ and the $\omega_j$ frames before it, and for offline recognisers it also holds the $\omega_j$ frames after $t$, truncated at the video boundaries in both cases. The window mean is
\begin{equation}
\bar{\mathbf{u}}_t^{(j)}=\frac{1}{|\mathcal{I}_t^{(j)}|}\sum_{s\in\mathcal{I}_t^{(j)}}\mathbf{u}_s.
\label{eq:mean}
\end{equation}
Interactions of different durations call for context at different scales, so we use several windows of increasing width. 
The same windows are used for every recogniser and dataset. The means capture accumulated multi-scale evidence without learning temporal parameters or explicitly modelling frame order within each window.

\paragraph{Step 3: input and target matrices.} The readout input $\bm{\xi}_t$ retains the current evidence $\mathbf{u}_t$ alongside its means, so that the readout can weight them separately
\begin{equation}
    \bm{\xi}_t=[\mathbf{u}_t;\bar{\mathbf{u}}_t^{(1)};\ldots;\bar{\mathbf{u}}_t^{(M)}]\in\mathbb{R}^{D},
    \label{eq:xi}
\end{equation}
with $D=(M+1)(d+K_f)$. Each dimension of $\bm{\xi}_t$ is standardised with its mean and standard deviation over the fitting frames, with the same statistics applied to validation and test frames. The frame-level target is $\tilde{\mathbf{y}}_t=[\mathbf{y}_t^\mathrm{IVT};\mathbf{y}_t^\mathrm{I};\mathbf{y}_t^\mathrm{V};\mathbf{y}_t^\mathrm{T}]\in\{0,1\}^{K}$, with $K=C+N_I+N_V+N_T$. Its component labels require no additional annotation, since a component is active whenever an active triplet contains it: $y^\mathrm{I}_{t,k}=\max_{c:\,i(c)=k}y^\mathrm{IVT}_{t,c}$, and analogously for verbs and targets.
We stack the standardised input vectors and their corresponding targets as rows of $\mathbf{Q}\in\mathbb{R}^{N\times D}$ and
$\mathbf{Y}\in\{0,1\}^{N\times K}$, respectively, where $N$ is
the number of fitting frames.

\paragraph{Step 4: closed-form readout.} We fit $\mathbf{W}\in\mathbb{R}^{D\times K}$ by ridge regression
\begin{equation}
    \min_{\mathbf{W}} \|\mathbf{Q}\mathbf{W}-\mathbf{Y}\|_F^2+\lambda\|\mathbf{W}\|_F^2,
    \label{eq:ridge}
\end{equation}
which has the closed-form solution
\begin{equation}
\mathbf{W}=(\mathbf{Q}^{\top}\mathbf{Q}+\lambda \mathbf{I})^{-1}\mathbf{Q}^{\top}\mathbf{Y}.
    \label{eq:cf}
\end{equation}
Within each cross-validation fold, the training videos are divided into fitting and validation videos. $\mathbf{W}$ is fitted on the fitting frames on which $f$ itself was trained, and $\lambda$ is selected by \apivt{} on the validation videos. Fitting $\mathbf{W}$ on out-of-fold predictions instead, as in stacked learning~\cite{wolpert1992stacked,cohen2005stacked}, does not improve results (\cref{sec:supp-crossfit}). 

\paragraph{Step 5: inference.} For a new frame, the readout gives one score for each of the $K$ targets,
\begin{equation}
    \hat{\mathbf{y}}_t = [\hat{\mathbf{y}}_t^\mathrm{IVT};\hat{\mathbf{y}}_t^\mathrm{I};\hat{\mathbf{y}}_t^\mathrm{V};\hat{\mathbf{y}}_t^\mathrm{T}]=\mathbf{W}^{\top}\bm{\xi}_t,
    \label{eq:inference}
\end{equation}
that is, triplet, instrument, verb, and target scores. A triplet $c$ that never occurs in the fitting videos has a zero column in $\mathbf{Y}$, so the readout assigns it a constant score. We replace the score of such a triplet by the product of its component scores, $
    \hat y^\mathrm{IVT}_{t,c}=\hat y^\mathrm{I}_{t,i(c)}\,\hat y^\mathrm{V}_{t,v(c)}\,\hat y^\mathrm{T}_{t,\tau(c)}
$.

\subsection{Adapting vision foundation models}
\label{sec:foundation}
We introduce a simple architecture for surgical triplet recognition that replaces the specialised heads in \cref{fig:X}a with a linear classifier. It combines a foundation vision encoder, the classifier, and STRIDE as shown in \cref{fig:X}b.

We train the framework in two stages. First, we adapt the foundation encoder $\phi_\theta$ with LoRA~\cite{hu2022lora} and jointly train the linear classifier head $h_\psi$. An image encoder processes the current frame, while a video encoder processes a clip associated with that frame. Given the pooled representation $\mathbf{z}_t$,
the linear classifier produces triplet and component logits
$\mathbf{a}_t=h_\psi(\mathbf{z}_t)$.%
We optimise the LoRA parameters and the classifier using weighted binary cross-entropy on the augmented targets $\tilde{\mathbf{y}}_t$ defined in \cref{sec:stride}.
In the second stage, we freeze the adapted encoder and the classifier, and fit STRIDE on their features $\mathbf{z}_t$ and logits $\mathbf{a}_t$ as described in \cref{sec:stride}.

As we show in \cref{sec:exp_encoders}, this simple architecture provides a strong baseline for surgical triplet recognition with both image and video encoders. The same two-stage procedure can be used to incorporate new foundation encoders as they become available.

\providecolor{dgrey}{gray}{0.6}\providecolor{ivt}{gray}{0.90}\providecolor{dup}{HTML}{2E8B57}\providecolor{ddown}{HTML}{D55E00}
\begin{table*}[t]\centering\footnotesize\setlength{\tabcolsep}{4pt}
\begin{tabular}{l@{\hspace{24pt}}rrr>{\columncolor{ivtbase}}r@{\hspace{24pt}}rrr>{\columncolor{ivtbase}}r@{\hspace{24pt}}rrr>{\columncolor{ivtbase}}r}
\toprule
 & \multicolumn{4}{c}{CholecT45} & \multicolumn{4}{c}{CholecT50} & \multicolumn{4}{c}{ProstaTD} \\
\cmidrule(lr){2-5}\cmidrule(lr){6-9}\cmidrule(lr){10-13}
Recogniser & AP$_\mathrm{I}$ & AP$_\mathrm{V}$ & AP$_{\mathrm{T}}$ & AP$_{\mathrm{IVT}}$ & AP$_\mathrm{I}$ & AP$_\mathrm{V}$ & AP$_{\mathrm{T}}$ & AP$_{\mathrm{IVT}}$ & AP$_\mathrm{I}$ & AP$_\mathrm{V}$ & AP$_{\mathrm{T}}$ & AP$_{\mathrm{IVT}}$ \\
\midrule
{\textbf{Online recognisers}} \\ \midrule
Tripnet~\cite{nwoye2020tripnet}  & 83.5 & 59.9 & 37.7 & 24.7\phantom{$^{\text{\scriptsize +0.0}}$} & 86.2 & 59.8 & 38.9 & 25.6\phantom{$^{\text{\scriptsize +0.0}}$} & 82.4 & 55.6 & 47.9 & 22.0\phantom{$^{\text{\scriptsize +0.0}}$} \\
\quad + STRIDE  & 90.6 & 65.8 & 43.4 & \cellcolor{ivtstride}32.1$^{\textcolor{dup}{\text{\scriptsize \textbf{+7.4}}}}$ & 90.2 & 63.8 & 43.5 & \cellcolor{ivtstride}30.9$^{\textcolor{dup}{\text{\scriptsize \textbf{+5.3}}}}$ & 85.2 & 61.9 & 54.7 & \cellcolor{ivtstride}28.9$^{\textcolor{dup}{\text{\scriptsize \textbf{+6.9}}}}$ \\
\bottomrule
Attention-Tripnet~\cite{nwoye2022rendezvous}  & 83.8 & 59.2 & 37.8 & 25.9\phantom{$^{\text{\scriptsize +0.0}}$} & 86.3 & 60.5 & 41.2 & 26.8\phantom{$^{\text{\scriptsize +0.0}}$} & 82.6 & 57.1 & 49.7 & 22.9\phantom{$^{\text{\scriptsize +0.0}}$} \\
\quad + STRIDE  & 88.8 & 65.1 & 43.4 & \cellcolor{ivtstride}31.7$^{\textcolor{dup}{\text{\scriptsize \textbf{+5.9}}}}$ & 89.9 & 64.8 & 46.2 & \cellcolor{ivtstride}31.3$^{\textcolor{dup}{\text{\scriptsize \textbf{+4.5}}}}$ & 86.5 & 63.4 & 55.8 & \cellcolor{ivtstride}29.5$^{\textcolor{dup}{\text{\scriptsize \textbf{+6.7}}}}$ \\
\bottomrule
Rendezvous~\cite{nwoye2022rendezvous}  & 86.2 & 61.4 & 39.4 & 29.4\phantom{$^{\text{\scriptsize +0.0}}$} & 87.2 & 60.6 & 41.0 & 29.0\phantom{$^{\text{\scriptsize +0.0}}$} & 80.6 & 55.9 & 49.1 & 23.0\phantom{$^{\text{\scriptsize +0.0}}$} \\
\quad + STRIDE  & 89.2 & 65.2 & 44.9 & \cellcolor{ivtstride}32.0$^{\textcolor{dup}{\text{\scriptsize \textbf{+2.6}}}}$ & 90.4 & 63.6 & 46.2 & \cellcolor{ivtstride}32.1$^{\textcolor{dup}{\text{\scriptsize \textbf{+3.1}}}}$ & 85.7 & 62.4 & 56.0 & \cellcolor{ivtstride}28.9$^{\textcolor{dup}{\text{\scriptsize \textbf{+6.0}}}}$ \\
\bottomrule
SDSwin~\cite{yamlahi2023sdswin}  & 90.4 & 68.2 & 48.3 & 36.6\phantom{$^{\text{\scriptsize +0.0}}$} & 89.6 & 66.1 & 47.6 & 35.2\phantom{$^{\text{\scriptsize +0.0}}$} & 92.0 & 70.6 & 64.3 & 37.6\phantom{$^{\text{\scriptsize +0.0}}$} \\
\quad + STRIDE  & 92.6 & 70.2 & 50.4 & \cellcolor{ivtstride}40.0$^{\textcolor{dup}{\text{\scriptsize \textbf{+3.4}}}}$ & 91.3 & 67.5 & 50.1 & \cellcolor{ivtstride}37.8$^{\textcolor{dup}{\text{\scriptsize \textbf{+2.7}}}}$ & 93.5 & 72.1 & 66.5 & \cellcolor{ivtstride}39.8$^{\textcolor{dup}{\text{\scriptsize \textbf{+2.1}}}}$ \\
\bottomrule
EndoViT~\cite{batic2024endovit}  & 84.2 & 61.8 & 41.1 & 28.7\phantom{$^{\text{\scriptsize +0.0}}$} & 83.7 & 59.7 & 40.1 & 28.0\phantom{$^{\text{\scriptsize +0.0}}$} & 87.7 & 63.2 & 57.5 & 30.9\phantom{$^{\text{\scriptsize +0.0}}$} \\
\quad + STRIDE  & 89.7 & 66.1 & 45.0 & \cellcolor{ivtstride}34.2$^{\textcolor{dup}{\text{\scriptsize \textbf{+5.5}}}}$ & 88.7 & 64.0 & 44.3 & \cellcolor{ivtstride}31.9$^{\textcolor{dup}{\text{\scriptsize \textbf{+3.9}}}}$ & 88.0 & 65.8 & 59.8 & \cellcolor{ivtstride}33.5$^{\textcolor{dup}{\text{\scriptsize \textbf{+2.6}}}}$ \\
\bottomrule
SPIRIT-S~\cite{sharma2026spirit}  & 94.7 & 69.1 & 47.5 & 38.6\phantom{$^{\text{\scriptsize +0.0}}$} & 94.2 & 67.0 & 48.7 & 39.4\phantom{$^{\text{\scriptsize +0.0}}$} & 95.0 & 72.5 & 65.1 & 40.0\phantom{$^{\text{\scriptsize +0.0}}$} \\
\quad + STRIDE  & 94.4 & 71.1 & 50.4 & \cellcolor{ivtstride}40.3$^{\textcolor{dup}{\text{\scriptsize \textbf{+1.7}}}}$ & 93.7 & 70.2 & 51.1 & \cellcolor{ivtstride}41.1$^{\textcolor{dup}{\text{\scriptsize \textbf{+1.7}}}}$ & 95.9 & 73.7 & 68.6 & \cellcolor{ivtstride}42.4$^{\textcolor{dup}{\text{\scriptsize \textbf{+2.4}}}}$ \\
\bottomrule
SPIRIT-TS~\cite{sharma2026spirit}  & 95.3 & 72.0 & 50.8 & 42.4\phantom{$^{\text{\scriptsize +0.0}}$} & 94.6 & 69.4 & 50.3 & 41.3\phantom{$^{\text{\scriptsize +0.0}}$} & 95.9 & 74.6 & 67.1 & 42.6\phantom{$^{\text{\scriptsize +0.0}}$} \\
\quad + STRIDE  & 94.7 & 72.8 & 51.9 & \cellcolor{ivtstride}42.2$^{\textcolor{dgrey}{\text{\scriptsize \textbf{$-$0.1}}}}$ & 93.8 & 70.9 & 52.1 & \cellcolor{ivtstride}41.2$^{\textcolor{dgrey}{\text{\scriptsize \textbf{$-$0.1}}}}$ & 96.1 & 74.3 & 69.5 & \cellcolor{ivtstride}43.1$^{\textcolor{dgrey}{\text{\scriptsize \textbf{+0.5}}}}$ \\
\specialrule{\heavyrulewidth}{\aboverulesep}{\belowrulesep}
{\textbf{Offline recognisers}} \\ \midrule
TERL~\cite{gui2024terl}  & 93.6 & 71.5 & 47.8 & {34.8}\phantom{$^{\text{\scriptsize +0.0}}$} & 93.7 & 70.1 & 48.7 & 34.4\phantom{$^{\text{\scriptsize +0.0}}$} & 93.7 & 70.8 & 63.7 & 36.8\phantom{$^{\text{\scriptsize +0.0}}$} \\
\quad + STRIDE  & 94.9 & 73.0 & 49.6 & \cellcolor{ivtstride}39.6$^{\textcolor{dup}{\text{\scriptsize \textbf{+4.8}}}}$ & 94.0 & 70.7 & 50.0 & \cellcolor{ivtstride}37.5$^{\textcolor{dup}{\text{\scriptsize \textbf{+3.1}}}}$ & 93.1 & 69.7 & 65.7 & \cellcolor{ivtstride}36.1$^{\textcolor{dgrey}{\text{\scriptsize \textbf{$-$0.6}}}}$ \\
\bottomrule
CurConMix+~\cite{jeon2026curconmixplus}  & 93.4 & 71.4 & 50.4 & 38.9\phantom{$^{\text{\scriptsize +0.0}}$} & 92.4 & 68.9 & 50.0 & 37.8\phantom{$^{\text{\scriptsize +0.0}}$} & 93.4 & 71.5 & 65.4 & 39.7\phantom{$^{\text{\scriptsize +0.0}}$} \\
\quad + STRIDE  & 93.9 & 73.0 & 52.2 & \cellcolor{ivtstride}42.7$^{\textcolor{dup}{\text{\scriptsize \textbf{+3.8}}}}$ & 92.8 & 70.7 & 52.7 & \cellcolor{ivtstride}41.3$^{\textcolor{dup}{\text{\scriptsize \textbf{+3.5}}}}$ & 94.2 & 72.2 & 67.6 & \cellcolor{ivtstride}40.5$^{\textcolor{dgrey}{\text{\scriptsize \textbf{+0.8}}}}$ \\
\bottomrule
\end{tabular}
\caption{\textbf{STRIDE as a post-hoc readout for existing surgical triplet recognisers.}
We report AP (\%) averaged over the five official cross-validation folds. Superscripts on AP$_{\mathrm{IVT}}$ show the change $\delta$ after adding STRIDE, with changes below the noise floor ($|\delta| < 1.62$) shown in grey. STRIDE improves AP$_{\mathrm{IVT}}$ in 24 of 27 recogniser--dataset pairs, with the largest average gains for framewise recognisers.}

\label{tab:table1}
\end{table*}

\begin{table*}[t]\centering\footnotesize\setlength{\tabcolsep}{4pt}
\begin{tabular}{l@{\hspace{24pt}}rrr>{\columncolor{ivtbase}}r@{\hspace{24pt}}rrr>{\columncolor{ivtbase}}r@{\hspace{24pt}}rrr>{\columncolor{ivtbase}}r}
\toprule
 & \multicolumn{4}{c}{CholecT45} & \multicolumn{4}{c}{CholecT50} & \multicolumn{4}{c}{ProstaTD} \\
\cmidrule(lr){2-5}\cmidrule(lr){6-9}\cmidrule(lr){10-13}
Encoder & AP$_\mathrm{I}$ & AP$_\mathrm{V}$ & AP$_\mathrm{T}$ & AP$_\mathrm{IVT}$ & AP$_\mathrm{I}$ & AP$_\mathrm{V}$ & AP$_\mathrm{T}$ & AP$_\mathrm{IVT}$ & AP$_\mathrm{I}$ & AP$_\mathrm{V}$ & AP$_\mathrm{T}$ & AP$_\mathrm{IVT}$ \\
\midrule
{\textbf{Image encoders (online)}} \\ \midrule
EndoViT~\cite{batic2024endovit} & 77.3 & 56.0 & 43.8 & 28.0\phantom{$^{\text{\scriptsize +0.0}}$} & 79.1 & 56.7 & 43.3 & 28.0\phantom{$^{\text{\scriptsize +0.0}}$} & 85.0 & 60.2 & 55.7 & 28.6\phantom{$^{\text{\scriptsize +0.0}}$} \\
\quad + STRIDE & 84.6 & 62.0 & 48.1 & \cellcolor{ivtstride}33.3$^{\textcolor{dup}{\text{\scriptsize \textbf{+5.4}}}}$ & 86.1 & 61.3 & 46.4 & \cellcolor{ivtstride}32.4$^{\textcolor{dup}{\text{\scriptsize \textbf{+4.3}}}}$ & 86.0 & 64.1 & 60.0 & \cellcolor{ivtstride}32.1$^{\textcolor{dup}{\text{\scriptsize \textbf{+3.5}}}}$ \\
\midrule
PeskaVLP~\cite{yuan2024peskavlp} & 78.6 & 58.4 & 41.0 & 26.2\phantom{$^{\text{\scriptsize +0.0}}$} & 78.8 & 57.6 & 40.9 & 25.4\phantom{$^{\text{\scriptsize +0.0}}$} & 82.9 & 59.4 & 55.1 & 28.5\phantom{$^{\text{\scriptsize +0.0}}$} \\
\quad + STRIDE & 88.4 & 65.2 & 45.7 & \cellcolor{ivtstride}33.3$^{\textcolor{dup}{\text{\scriptsize \textbf{+7.2}}}}$ & 88.4 & 63.5 & 46.6 & \cellcolor{ivtstride}32.5$^{\textcolor{dup}{\text{\scriptsize \textbf{+7.1}}}}$ & 86.6 & 63.9 & 61.7 & \cellcolor{ivtstride}31.4$^{\textcolor{dup}{\text{\scriptsize \textbf{+2.9}}}}$ \\
\midrule
LemonFM~\cite{che2026lemon} & 87.4 & 66.6 & 48.4 & 34.1\phantom{$^{\text{\scriptsize +0.0}}$} & 87.5 & 66.1 & 48.0 & 33.9\phantom{$^{\text{\scriptsize +0.0}}$} & 91.7 & 68.9 & 64.9 & 37.1\phantom{$^{\text{\scriptsize +0.0}}$} \\
\quad + STRIDE & 92.0 & 68.6 & 50.2 & \cellcolor{ivtstride}37.3$^{\textcolor{dup}{\text{\scriptsize \textbf{+3.2}}}}$ & 90.9 & 66.5 & 49.8 & \cellcolor{ivtstride}35.5$^{\textcolor{dup}{\text{\scriptsize \textbf{+1.6}}}}$ & 91.3 & 70.3 & 66.8 & \cellcolor{ivtstride}38.2$^{\textcolor{dgrey}{\text{\scriptsize \textbf{+1.1}}}}$ \\
\midrule
SigLIP2-L~\cite{tschannen2025siglip2} & 88.7 & 66.2 & 46.8 & 33.4\phantom{$^{\text{\scriptsize +0.0}}$} & 88.6 & 65.5 & 46.8 & 31.9\phantom{$^{\text{\scriptsize +0.0}}$} & 91.2 & 68.5 & 66.2 & 37.6\phantom{$^{\text{\scriptsize +0.0}}$} \\
\quad + STRIDE & 92.6 & 69.4 & 49.7 & \cellcolor{ivtstride}39.0$^{\textcolor{dup}{\text{\scriptsize \textbf{+5.5}}}}$ & 91.9 & 68.2 & 50.3 & \cellcolor{ivtstride}36.2$^{\textcolor{dup}{\text{\scriptsize \textbf{+4.3}}}}$ & 92.5 & 70.0 & 67.3 & \cellcolor{ivtstride}39.0$^{\textcolor{dgrey}{\text{\scriptsize \textbf{+1.4}}}}$ \\
\midrule
DINOv3~\cite{simeoni2025dinov3} & 89.4 & 66.9 & 47.4 & 33.5\phantom{$^{\text{\scriptsize +0.0}}$} & 88.9 & 66.0 & 47.8 & 33.0\phantom{$^{\text{\scriptsize +0.0}}$} & 91.1 & 69.0 & 64.5 & 37.7\phantom{$^{\text{\scriptsize +0.0}}$} \\
\quad + STRIDE & 92.5 & 70.2 & 50.7 & \cellcolor{ivtstride}37.3$^{\textcolor{dup}{\text{\scriptsize \textbf{+3.8}}}}$ & 91.7 & 68.8 & 51.6 & \cellcolor{ivtstride}36.8$^{\textcolor{dup}{\text{\scriptsize \textbf{+3.7}}}}$ & 91.6 & 71.2 & 67.4 & \cellcolor{ivtstride}39.4$^{\textcolor{dup}{\text{\scriptsize \textbf{+1.7}}}}$ \\
\midrule
SurgLIME~\cite{che2026surglime} & 87.7 & 66.2 & 48.3 & 34.1\phantom{$^{\text{\scriptsize +0.0}}$} & 87.4 & 65.5 & 47.6 & 33.3\phantom{$^{\text{\scriptsize +0.0}}$} & 91.9 & 70.3 & 66.2 & 38.5\phantom{$^{\text{\scriptsize +0.0}}$} \\
\quad + STRIDE & 91.8 & 68.6 & 50.1 & \cellcolor{ivtstride}38.4$^{\textcolor{dup}{\text{\scriptsize \textbf{+4.3}}}}$ & 91.2 & 67.6 & 50.4 & \cellcolor{ivtstride}37.5$^{\textcolor{dup}{\text{\scriptsize \textbf{+4.2}}}}$ & 91.9 & 72.2 & 68.0 & \cellcolor{ivtstride}41.0$^{\textcolor{dup}{\text{\scriptsize \textbf{+2.4}}}}$ \\
\midrule
SurgeNetXL~\cite{jaspers2025surgenetxl} & 89.2 & 67.2 & 49.6 & 35.1\phantom{$^{\text{\scriptsize +0.0}}$} & 89.1 & 66.1 & 50.3 & 34.4\phantom{$^{\text{\scriptsize +0.0}}$} & 93.5 & 69.2 & 66.0 & 38.0\phantom{$^{\text{\scriptsize +0.0}}$} \\
\quad + STRIDE & 92.1 & 71.0 & 52.5 & \cellcolor{ivtstride}38.9$^{\textcolor{dup}{\text{\scriptsize \textbf{+3.9}}}}$ & 91.7 & 69.3 & 53.1 & \cellcolor{ivtstride}37.7$^{\textcolor{dup}{\text{\scriptsize \textbf{+3.3}}}}$ & 94.3 & 72.1 & 68.7 & \cellcolor{ivtstride}41.3$^{\textcolor{dup}{\text{\scriptsize \textbf{+3.2}}}}$ \\
\midrule
PL-Stitch~\cite{che2026plstitch} & 87.8 & 66.8 & 48.2 & 34.7\phantom{$^{\text{\scriptsize +0.0}}$} & 87.0 & 65.0 & 47.4 & 33.9\phantom{$^{\text{\scriptsize +0.0}}$} & 91.8 & 70.3 & 66.2 & 38.8\phantom{$^{\text{\scriptsize +0.0}}$} \\
\quad + STRIDE & 92.0 & 69.8 & 50.8 & \cellcolor{ivtstride}39.1$^{\textcolor{dup}{\text{\scriptsize \textbf{+4.5}}}}$ & 91.3 & 68.3 & 51.0 & \cellcolor{ivtstride}38.4$^{\textcolor{dup}{\text{\scriptsize \textbf{+4.4}}}}$ & 92.3 & 72.3 & 67.6 & \cellcolor{ivtstride}41.2$^{\textcolor{dup}{\text{\scriptsize \textbf{+2.4}}}}$ \\
\midrule
{\textbf{Video encoders (offline)}} \\ \midrule
VideoMAE V2~\cite{wang2023videomaev2} & 91.0 & 69.8 & 50.1 & 38.1\phantom{$^{\text{\scriptsize +0.0}}$} & 91.0 & 68.3 & 50.8 & 37.8\phantom{$^{\text{\scriptsize +0.0}}$} & 88.6 & 74.1 & 66.0 & 37.8\phantom{$^{\text{\scriptsize +0.0}}$} \\
\quad + STRIDE & 90.9 & 70.3 & 51.4 & \cellcolor{ivtstride}40.5$^{\textcolor{dup}{\text{\scriptsize \textbf{+2.3}}}}$ & 91.5 & 69.2 & 51.8 & \cellcolor{ivtstride}38.9$^{\textcolor{dgrey}{\text{\scriptsize \textbf{+1.1}}}}$ & 87.5 & 74.7 & 67.2 & \cellcolor{ivtstride}38.5$^{\textcolor{dgrey}{\text{\scriptsize \textbf{+0.7}}}}$ \\
\midrule
V-JEPA 2~\cite{assran2025vjepa2} & 90.3 & 72.7 & 52.5 & 38.5\phantom{$^{\text{\scriptsize +0.0}}$} & 90.5 & 70.8 & 51.3 & 37.6\phantom{$^{\text{\scriptsize +0.0}}$} & 88.6 & 75.9 & 68.8 & 40.8\phantom{$^{\text{\scriptsize +0.0}}$} \\
\quad + STRIDE & 92.2 & 74.2 & 53.5 & \cellcolor{ivtstride}43.5$^{\textcolor{dup}{\text{\scriptsize \textbf{+5.1}}}}$ & 92.2 & 72.4 & 53.7 & \cellcolor{ivtstride}41.6$^{\textcolor{dup}{\text{\scriptsize \textbf{+3.9}}}}$ & 88.8 & 77.8 & 70.6 & \cellcolor{ivtstride}41.4$^{\textcolor{dgrey}{\text{\scriptsize \textbf{+0.5}}}}$ \\
\midrule
InternVideo2~\cite{wang2024internvideo2} & 91.8 & 72.0 & 51.7 & 40.2\phantom{$^{\text{\scriptsize +0.0}}$} & 92.6 & 70.9 & 51.9 & 39.6\phantom{$^{\text{\scriptsize +0.0}}$} & 90.5 & 74.7 & 70.3 & 40.7\phantom{$^{\text{\scriptsize +0.0}}$} \\
\quad + STRIDE & 93.0 & 72.1 & 52.3 & \cellcolor{ivtstride}42.8$^{\textcolor{dup}{\text{\scriptsize \textbf{+2.6}}}}$ & 93.1 & 71.6 & 52.6 & \cellcolor{ivtstride}41.7$^{\textcolor{dup}{\text{\scriptsize \textbf{+2.1}}}}$ & 90.3 & 75.5 & 70.8 & \cellcolor{ivtstride}40.1$^{\textcolor{dgrey}{\text{\scriptsize \textbf{$-$0.6}}}}$ \\
\midrule
TrackMAE~\cite{vandeghen2026trackmae} & 93.0 & 75.2 & 53.7 & 41.9\phantom{$^{\text{\scriptsize +0.0}}$} & 93.2 & 72.4 & 54.3 & 40.7\phantom{$^{\text{\scriptsize +0.0}}$} & 92.5 & 78.5 & 71.6 & 44.8\phantom{$^{\text{\scriptsize +0.0}}$} \\
\quad + STRIDE & 93.2 & 75.1 & 54.0 & \cellcolor{ivtstride}45.1$^{\textcolor{dup}{\text{\scriptsize \textbf{+3.2}}}}$ & 93.4 & 72.8 & 54.1 & \cellcolor{ivtstride}43.0$^{\textcolor{dup}{\text{\scriptsize \textbf{+2.4}}}}$ & 92.2 & 79.4 & 72.1 & \cellcolor{ivtstride}44.6$^{\textcolor{dgrey}{\text{\scriptsize \textbf{$-$0.2}}}}$ \\
\midrule
SurgRec-JEPA$^\dagger$~\cite{lu2026surgrec} & 93.6 & 78.3 & 58.0 & 46.2\phantom{$^{\text{\scriptsize +0.0}}$} & 93.5 & 75.8 & 57.7 & 44.4\phantom{$^{\text{\scriptsize +0.0}}$} & 92.7 & 80.2 & 73.2 & 47.0\phantom{$^{\text{\scriptsize +0.0}}$} \\
\quad + STRIDE & 94.1 & 79.3 & 58.8 & \cellcolor{ivtstride}50.7$^{\textcolor{dup}{\text{\scriptsize \textbf{+4.5}}}}$ & 94.1 & 77.4 & 59.2 & \cellcolor{ivtstride}48.1$^{\textcolor{dup}{\text{\scriptsize \textbf{+3.7}}}}$ & 93.5 & 81.7 & 74.5 & \cellcolor{ivtstride}48.3$^{\textcolor{dgrey}{\text{\scriptsize \textbf{+1.4}}}}$ \\
\bottomrule
\end{tabular}
\caption{STRIDE on foundation encoders adapted with LoRA and a linear classifier. Mean AP (\%) over the official cross-validation folds. Superscript numbers: (LoRA + STRIDE) $-$ LoRA, grey for $|\delta|$$\,<\,$$1.62$, i.e. noise floor. $^\dagger\!$ Pre-training contains all CholecT50 test videos.}
\label{tab:table2}
\end{table*}

\section{Experiments}
\label{sec:experiments}

We evaluate STRIDE on three public surgical triplet recognition datasets (CholecT45~\cite{nwoye2022splits}, CholecT50~\cite{nwoye2022rendezvous}, and ProstaTD~\cite{chen2026prostatd}). We first apply STRIDE as a post-hoc readout to nine existing recognisers (\cref{sec:exp_methods}) and evaluate our foundation-model architecture (\cref{sec:foundation}) with thirteen vision encoders (\cref{sec:exp_encoders}). We then ablate the components of STRIDE (\cref{sec:ablations}), examine how it changes predictions over time (\cref{sec:temporal_analysis}), and analyse how its gains vary with triplet frequency and duration (\cref{sec:freq}).

\subsection{Implementation details}
\label{sec:impl}

\paragraph{Datasets.}
CholecT50~\cite{nwoye2022rendezvous} comprises 50 laparoscopic cholecystectomy videos with 100.9K frames, annotated with 100 triplets over 6 instruments, 10 verbs, and 15 targets. CholecT45~\cite{nwoye2022splits} is its public subset of 45 videos and 90.5K frames. ProstaTD~\cite{chen2026prostatd} comprises 21 robot-assisted prostatectomy videos from three centres, with 71.8K frames and 89 triplets over 7 instruments, 10 verbs and 10 targets. All videos are sampled at 1\,fps. We use the official five-fold cross-validation splits for each dataset.

\paragraph{Evaluation.}
Following prior work~\cite{nwoye2022splits}, we report average precision (AP), computed with \texttt{ivtmetrics}, the official evaluation library for surgical triplet benchmarks.
AP is computed per class within each video and averaged over videos. \apivt{} is the mean over the triplet classes, and AP$_\mathrm{I}$, AP$_\mathrm{V}$ and AP$_\mathrm{T}$ are the mean AP over the instrument, verb, and target classes. We report the mean over the five folds and additionally average over multiple seeds where possible. We set a noise floor of 1.62 \apivt{}, based on the standard deviation of paired five-fold differences across three seeds. 
We call an improvement reliable if it exceeds this floor and is positive on at least four folds. Paired 95\% confidence intervals are given in the supplement~(\cref{sec:supp-stats}). Both rules lead to the same conclusion for every pair in~\Cref{tab:table1}.

\paragraph{STRIDE.}
We use $M=3$ windows with $\omega_j\in\{2,8,32\}$\,s.
To select $\lambda$ we hold out five training videos per fold on CholecT45/50 and two on ProstaTD, which are excluded from fitting the readout. The grid has 15 log-spaced values in $[10^{1}, 10^{8}]$, or seven in $[10^{3}, 3\cdot10^{6}]$ for offline methods and video encoders.
STRIDE is fitted on a CPU.

\paragraph{Adaptation.}
We adapt each foundation encoder using LoRA~\cite{hu2022lora} with rank 16 and $\alpha=32$, applied to the attention projections. We train for four epochs using AdamW, a learning rate of $5\cdot10^{-4}$ and a batch size of 64, and retain the final epoch. Image encoders take the current frame. Video encoders take a 16-frame clip at 1\,fps centred on the current frame, and InternVideo2 takes 8 frames at a stride of two. Boundary frames are repeated at the start and end of each video. 
The validation videos are excluded from adaptation. We use NVIDIA H100 GPUs for adaptation.

\subsection{Plug-and-play with existing architectures}
\label{sec:exp_methods}

\paragraph{Evaluated architectures.}
We evaluate STRIDE on nine existing recognisers: Tripnet~\cite{nwoye2020tripnet}, Attention-Tripnet and Rendezvous~\cite{nwoye2022rendezvous}, SDSwin~\cite{yamlahi2023sdswin}, EndoViT~\cite{batic2024endovit}, SPIRIT-S and SPIRIT-TS~\cite{sharma2026spirit}, TERL~\cite{gui2024terl} and CurConMix+~\cite{jeon2026curconmixplus}. These methods operate in online settings, except TERL and CurConMix+, which also use future frames. 
We use released checkpoints trained on the official cross-validation folds where available and otherwise retrain each recogniser on those folds with the authors’ code and settings.
Since SPIRIT has no public implementation, we re-implement it following the original description~(\cref{sec:supp-recipes}). All methods are kept fixed when fitting STRIDE. We use causal windows for online methods and centred windows for offline methods.

\paragraph{Results.}
\Cref{tab:table1} reports the results across all datasets. STRIDE improves \apivt{} in 24 of the 27 method--dataset pairs, 22 of them beyond the noise floor. The gains are largest for methods that predict from single frames, 4.6 points on average for Tripnet, Attention-Tripnet, Rendezvous, SDSwin and EndoViT, against 1.8 for the four methods with their own temporal module. 
This suggests that the readout is particularly useful when the base prediction comes from a single frame. 
Of the SPIRIT variants, STRIDE improves SPIRIT-S by 1.7--2.4 \apivt{} points across the three datasets, while the changes for SPIRIT-TS, which uses teacher--student distillation, remain within the noise floor.

The two offline methods improve on the Cholec datasets but change little on ProstaTD, whereas STRIDE improves every online method there. ProstaTD interactions are shorter (median 4\,s against 9\,s, \cref{sec:supp-labels}), so a bidirectional temporal module plausibly covers most of the useful context, whereas in the standard online setting STRIDE supplies past context the recogniser does not exploit.

STRIDE improves AP$_\mathrm{T}$ in all 27 pairs and AP$_\mathrm{V}$ in all but two. AP$_\mathrm{I}$ decreases slightly for SPIRIT on the Cholec datasets and for TERL on ProstaTD, where it already exceeds 93 AP. The target is the weakest component of every recogniser, and on average it is the component that STRIDE improves most, including in the pairs where \apivt{} does not improve. 
A triplet is correct only if its instrument, verb and target are all recognised and paired correctly. Since AP$_\mathrm{I}$ already exceeds 80 for every method, the verb and target leave the most room for improvement. 
Targets are anatomical structures that are often partly covered by the instrument acting on them, so their identity is plausibly easier to establish from preceding frames than from the current one.

In summary, STRIDE can act as a simple post-hoc plug-and-play module for existing models and improve their performance in both online and offline settings. 

\subsection{Adapting vision foundation models}
\label{sec:exp_encoders}

\paragraph{Encoders.}
We evaluate eight image encoders~\cite{che2026lemon,jaspers2025surgenetxl,batic2024endovit,tschannen2025siglip2,simeoni2025dinov3,che2026surglime,yuan2024peskavlp,che2026plstitch} online and five video encoders~\cite{vandeghen2026trackmae,wang2023videomaev2,assran2025vjepa2,wang2024internvideo2,lu2026surgrec} offline. Each encoder is adapted with LoRA and a linear classifier (\cref{sec:impl}). We then freeze both and fit STRIDE on the resulting features and logits. SurgRec-JEPA was pre-trained on data that include CholecT50 test videos, so we report it but exclude it from comparisons.

\paragraph{Results.}
\Cref{tab:table2} compares each adapted encoder with and without STRIDE on the three datasets. STRIDE improves \apivt{} in 37 of the 39 encoder--dataset pairs.

STRIDE improves all eight image encoders on every dataset, by 2.3 to 4.7 points on average and by up to 7.2 points. On the Cholec datasets, general-purpose and surgical encoders gain by a similar margin, whereas on ProstaTD the gain is about 70\% larger for surgical encoders. The gain also holds for the strongest encoders. PL-Stitch, which gives the best online results on both Cholec datasets with 39.1 and 38.4, still gains 4.5 and 4.4 points. With STRIDE, the general-purpose SigLIP2-L reaches 39.0 on CholecT45, on par with the best surgical encoder.

Video encoders already see a clip of surrounding frames, and their gain is smaller, 3.3 and 2.4 points on average on Cholec. TrackMAE is the strongest encoder offline on all three datasets, and reaches 45.1, 43.0 and 44.6 with STRIDE. On ProstaTD, the offline video encoders gain little in \apivt{}, although AP$_\mathrm{V}$ and AP$_\mathrm{T}$ still improve for all of them. This matches the offline methods in \cref{tab:table1}, whose bidirectional context already covers the shorter interactions of this dataset. The smaller gain of video encoders is not due to the offline setting, as image encoders still gain more when both are evaluated offline~(\cref{sec:supp-settings}). STRIDE therefore helps most where the encoder has the least temporal context. In summary, a foundation encoder with a linear classifier and STRIDE forms a simple pipeline that approaches specialised methods~(\cref{tab:table1}) in the online setting and exceeds them offline.

\subsection{Ablations}
\label{sec:ablations}

We ablate three design choices of STRIDE: which outputs of the recogniser it reads, which temporal scales it averages over, and how the readout is fitted. The baseline row $f$ denotes the recognisers without STRIDE. Unless stated otherwise, we average AP across the online methods in \Cref{tab:table1}.

\paragraph{What outputs of $f$ should STRIDE read?}
\Cref{tab:abl_input} compares readout inputs with and without temporal windows. Without windows ($M=0$), the features $\mathbf{z}_t$ fall below the original recogniser $f$, whereas the concatenated inputs improve on it. 
With windows, every input improves and $\mathbf{u}_t$ performs best. The probabilities $\mathbf{p}_t$ neither match the logits $\mathbf{a}_t$ nor add to them.
Features supply visual evidence and logits supply class-specific predictions. Features gain most from the windows, so the temporal context of visual evidence carries information the recogniser’s own scores lack.
\begin{table}[htbp]\centering\footnotesize\setlength{\tabcolsep}{8pt}
\begin{tabular}{lcccc}
\toprule
 & \multicolumn{2}{c}{CholecT50} & \multicolumn{2}{c}{ProstaTD} \\
\cmidrule(lr){2-3}\cmidrule(lr){4-5}
Input & $M=0$ & $M=3$ & $M=0$ & $M=3$ \\
\midrule
$\mathbf{z}_t$ & 30.3 & 33.7 & 31.1 & 33.9 \\
$\mathbf{p}_t$ & 31.5 & 33.3 & 31.4 & 33.3 \\
$\mathbf{a}_t$ & 32.6 & 34.6 & 32.0 & 33.9 \\
\midrule
$[\mathbf{z}_t;\mathbf{p}_t]$ & 32.6 & 34.4 & 32.7 & 34.8 \\
\rowcolor{gray!10}$[\mathbf{z}_t;\mathbf{a}_t]$ & 33.0 & {35.2} & 33.1 & {35.2} \\
$[\mathbf{z}_t;\mathbf{p}_t;\mathbf{a}_t]$ & 33.3 & 34.8 & 33.0 & 35.0 \\
\midrule
$f$ (baseline) & \multicolumn{2}{c}{32.2} & \multicolumn{2}{c}{31.3} \\
\bottomrule
\end{tabular}
\caption{Readout input, \apivt{}: features $\mathbf{z}_t$, scores $\mathbf{p}_t$ and logits $\mathbf{a}_t$ of $f$, without windows ($M=0$) and with $\omega_j\in\{2,8,32\}$\,s.}
\label{tab:abl_input}
\end{table}

\paragraph{Does the choice of temporal scales matter?}
Temporal context matters more than the exact scales. A single window of 2\,s already improves on $f$ by 2.4 and 2.9 points~(\cref{tab:abl_windows}). Combining a short and a long scale generally helps, consistent with interactions of different durations, whereas each further scale changes \apivt{} by at most 0.2. A single fixed set of scales therefore suffices for all recognisers.

\paragraph{Trained or closed-form?}
\Cref{tab:abl_readout} compares the closed-form ridge readout with iteratively trained alternatives. 
MS-TCN++ receives $\mathbf{u}_t$ rather than $\bm{\xi}_t$, since it models temporal context itself through dilated convolutions. All readouts improve on the baseline $f$. Ridge exceeds MS-TCN++ and matches the MLP, while L2-regularised logistic regression exceeds ridge by less than the noise floor. Overall, the window means account for most of the gain, and the closed-form solution gives up only a small margin to an iterative fit.

\begin{table}[htbp]
\centering
\footnotesize
\setlength{\tabcolsep}{2pt}
\resizebox{\linewidth}{!}{%
\begin{tabular}{clccc>{\columncolor{ivtbase}}c@{\hspace{5pt}}ccc>{\columncolor{ivtbase}}c}
\toprule
 & & \multicolumn{4}{c}{CholecT50} & \multicolumn{4}{c}{ProstaTD} \\
\cmidrule(lr){3-6}\cmidrule(lr){7-10}
$M$ & $\{\omega_j\}$ (s) & AP$_\mathrm{I}$ & AP$_\mathrm{V}$ & AP$_\mathrm{T}$ & AP$_\mathrm{IVT}$ & AP$_\mathrm{I}$ & AP$_\mathrm{V}$ & AP$_\mathrm{T}$ & AP$_\mathrm{IVT}$ \\
\midrule 
1 & 2 & 90.9 & 66.0 & 46.5 & 34.6 & 89.8 & 67.3 & 60.9 & 34.2 \\
2 & 2, 32 & 91.0 & 66.1 & 47.4 & 35.1 & 90.1 & 67.6 & 61.5 & 35.0 \\
3 & 2, 8, 32 & 91.1 & 66.4 & 47.6 & \cellcolor{ivtstride}35.2 & 90.1 & 67.7 & 61.6 & \cellcolor{ivtstride}35.2 \\
4 & 2, 8, 32, 128 & 91.0 & 66.2 & 47.5 & 35.0 & 90.2 & 67.7 & 61.5 & 35.3 \\
\midrule
$f$ & (baseline) & 88.8 & 63.3 & 44.0 & 32.2 & 88.0 & 64.2 & 57.2 & 31.3 \\
\bottomrule
\end{tabular}}
\caption{Representative temporal-window configurations on $u_t$, evaluated using \apivt{}. The full window ablation is provided in the supplement~(\cref{sec:supp-window}).}
\label{tab:abl_windows}
\end{table}
\providecolor{ivt}{gray}{0.90}
\begin{table}[htbp]\centering\footnotesize\setlength{\tabcolsep}{2pt}
\resizebox{\linewidth}{!}{%
\begin{tabular}{lccc>{\columncolor{ivtbase}}c@{\hspace{5pt}}ccc>{\columncolor{ivtbase}}c}
\toprule
 & \multicolumn{4}{c}{CholecT50} & \multicolumn{4}{c}{ProstaTD} \\
\cmidrule(lr){2-5}\cmidrule(lr){6-9}
Readout & AP$_\mathrm{I}$ & AP$_\mathrm{V}$ & AP$_\mathrm{T}$ & AP$_\mathrm{IVT}$ & AP$_\mathrm{I}$ & AP$_\mathrm{V}$ & AP$_\mathrm{T}$ & AP$_\mathrm{IVT}$ \\
\midrule
MS-TCN++ & 90.6 & 65.8 & 46.9 & 36.5 & 92.4 & 70.5 & 63.7 & 38.0 \\
\midrule
MLP & 92.2 & 67.3 & 47.9 & 37.9 & 93.4 & 71.3 & 64.4 & 39.1 \\
Logistic & 92.5 & 68.5 & 50.2 & 39.3 & 93.6 & 72.0 & 65.9 & 40.9 \\
Ridge & 91.9 & 68.1 & 49.4 & \cellcolor{ivtstride}38.0 & 93.4 & 71.5 & 66.1 & \cellcolor{ivtstride}39.7 \\
\midrule
$f$ (baseline) & 90.5 & 65.6 & 46.7 & 36.0 & 92.7 & 70.2 & 63.5 & 37.8 \\
\bottomrule
\end{tabular}}

\caption{Causal readouts on a fixed recogniser $f$. MS-TCN++ is trained on $\mathbf{u}_t$ and the rest on $\bm{\xi}_t$. MLP and logistic regression are trained iteratively, ridge is solved in closed form. $f$ is averaged over SDSwin, EndoViT, SPIRIT-S and SPIRIT-TS.}
\label{tab:abl_readout}
\end{table}

\subsection{Temporal behaviour}
\label{sec:temporal_analysis}
\begin{figure}[t]
    \centering
\includegraphics[width=\linewidth]{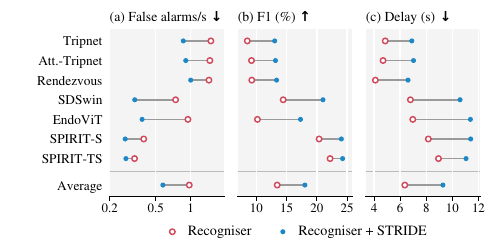}
    \caption{Segment-level metrics on CholecT50 for the online recognisers, without and with STRIDE, at matched recall 0.5.}
    \label{fig:M}
\end{figure}
\begin{figure*}[t]
    \centering    \includegraphics[width=\textwidth]{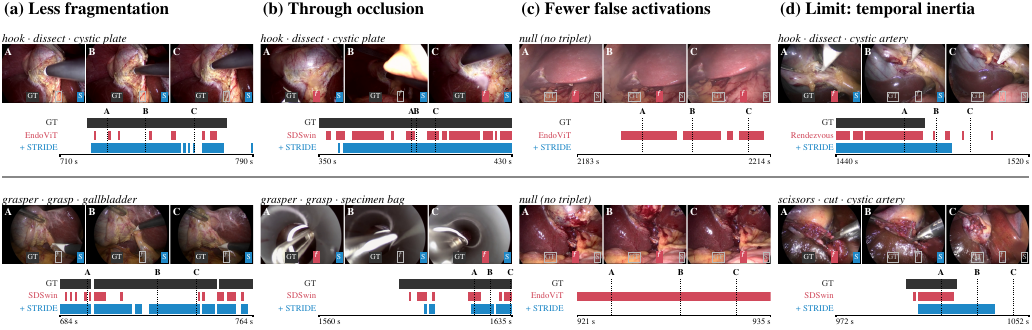}
    \caption{CholecT50 examples: labels (GT) and decisions of the recogniser $f$ and of $f$ + STRIDE (S) for the named triplet, at matched recall 0.5. A--C mark the frames shown; a filled tag means active. \textbf{(a)} STRIDE keeps fragmented predictions continuous. \textbf{(b)} View obstructed by smoke, glare or a specimen bag. \textbf{(c)} No triplet is active while $f$ predicts one. \textbf{(d)} Limit: STRIDE stays active after the labelled segment ends.}
    \label{fig:Z}
\end{figure*}
\paragraph{Segment-level metrics.}
\Cref{fig:M} compares seven online recognisers on CholecT50, with and without STRIDE, at matched recall 0.5. Averaged over recognisers, STRIDE reduces false alarms from 0.98 to 0.58 per second~(41\%), and raises segment F1 from 13.5\% to 18.0\%. Mean onset delay grows from 6.4 to 9.3\,s. All three changes have the same direction for every recogniser in all five folds. False alarms are summed over classes and normalised by video duration, and segment F1 uses an IoU threshold of 0.1.

\paragraph{Qualitative examples.}
\Cref{fig:Z} shows typical sources of these changes. STRIDE keeps predictions continuous within labelled segments~(a), retains them through smoke, glare or occlusion~(b), and suppresses some false activations of the recogniser~(c). It also introduces inertia~(d). For labelled segments of at least 10\,s, the fraction with no activation in the first 30\,s rises from 12.2\% to 19.6\%, and activation after a segment ends grows from 1.0 to 1.6\,s. STRIDE therefore trades responsiveness at interaction boundaries for fewer false alarms and more continuous segments. 

\begin{figure}[t]
\centering
\includegraphics[width=\linewidth]{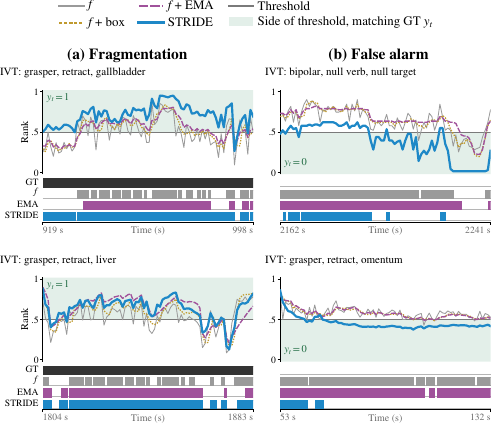}
\caption{CholecT50 examples. Rank-normalised scores place each line's threshold at a recall of 0.5. \textbf{(a)} Smoothing and STRIDE reduce fragmentation. \textbf{(b)} Smoothing retains erroneous activations of $f$, whereas STRIDE activates less often.}
\label{fig:smooth}
\end{figure}
\paragraph{STRIDE or smoothing?}
We compare STRIDE with a causal box filter and an exponential moving average (EMA) on the scores of $f$, each tuned on the validation videos for every recogniser. On CholecT50, the box filter and EMA improve \apivt{} by 0.9 and 1.9 points, and STRIDE by 3.0 points. On ProstaTD, the gains are 0.2, 1.5 and 3.9 points (\cref{sec:supp-smoothing}). A smoother can only change the timing of the recogniser's own activations. It reduces fragmentation (\cref{fig:smooth}a) but retains the erroneous activations (\cref{fig:smooth}b). STRIDE also reads the visual features, and can suppress an activation that they do not support. Accordingly, STRIDE raises AP$_\mathrm{V}$ and AP$_\mathrm{T}$ by about three times as much as EMA on both datasets.

\subsection{Class frequency and duration}
\label{sec:freq}
\Cref{tab:abl_freq} groups the triplet classes into thirds, first by their number of training frames and then by the median length of their labelled segments. STRIDE improves \apivt{} in every group. The gap between frequent and rare classes remains, however, so STRIDE does not resolve the class imbalance. Across triplet duration, the gain grows with the length of the interaction. 
Short interactions nevertheless still gain in performance despite their later onsets~(\cref{sec:temporal_analysis}). 
\newcommand{\hl}[1]{{\setlength{\fboxsep}{1.0pt}\smash{\colorbox{blue!10}{\vphantom{0}#1}}}}
\begin{table}[hbtp]\centering\footnotesize\setlength{\tabcolsep}{3.5pt}
\begin{tabular}{lcccc}
\toprule
Classes by & Lower third & Middle third & Upper third & All \\
\midrule
Frequency & 20.3$\,\to\,$\hl{23.7} & 28.4$\,\to\,$\hl{31.8} & 50.7$\,\to\,$\hl{53.9} & 31.9$\,\to\,$\hl{35.5} \\
Duration & 24.2$\,\to\,$\hl{26.9} & 36.1$\,\to\,$\hl{39.8} & 37.6$\,\to\,$\hl{42.0} & 31.9$\,\to\,$\hl{35.5} \\
\bottomrule
\end{tabular}
\caption{\apivt{} before $\to$ after STRIDE, averaged over the online recognisers and three datasets. Classes are split by training frames (rare to frequent) and by segment length (short to long).}
\label{tab:abl_freq}
\end{table}

\section{Conclusion}
\label{sec:conclusion}
We introduced \STRIDE{}, a closed-form temporal readout that uses visual features and logits from a trained surgical triplet recogniser. A fixed set of temporal scales improves recognition across architectures and datasets, with the largest gains for framewise models. The same readout, when paired with an adapted foundation encoder and a linear classifier, provides a competitive online baseline and outperforms the specialised offline recognizers.
The main limitation of STRIDE is temporal inertia, since it responds later at interaction boundaries. It cannot distinguish an ended interaction from one hidden by temporary obstruction. This cost is confined to boundaries, and \apivt{} still improves for the shortest interactions. The segment-level evaluation is also diagnostic only, as its operating point is set on test labels. Segment metrics complement frame-level AP but do not fully capture temporal localisation.
STRIDE requires no retraining, can be applied directly to stronger emerging encoders and offers a simple baseline for future triplet recognition methods.

\clearpage
\section*{LLM Usage Statement}
The authors used Anthropic Claude for reviewer style feedback and refinement of author-produced text and figures. AI agents were used for coding tasks. The authors take full responsibility for the paper.
{\small
\bibliographystyle{ieeenat_fullname}
\bibliography{ref}
}

 \clearpage
\setcounter{page}{1}
\maketitlesupplementary
\setcounter{section}{0}
\renewcommand{\thesection}{\Alph{section}}
\nocite{schuirmann1987tost}

\section{Method details}
\label{sec:supp-method}

\Cref{alg:stride} summarises \cref{sec:stride} for one cross-validation fold. The windows are causal for online and centred for offline recognisers. A single eigendecomposition of $\mathbf{Q}_\mathcal{F}^{\top}\mathbf{Q}_\mathcal{F}$ yields the ridge solution for every $\lambda$ in the grid~(\cref{sec:impl}), and we keep the $\lambda$ with the highest \apivt{} on the validation videos.

\begin{algorithm}[htbp]
\caption{STRIDE for one cross-validation fold.}
\label{alg:stride}
\footnotesize
\begin{algorithmic}[1]
\Require recogniser $f$, fitting videos $\mathcal{F}$, validation videos $\mathcal{V}$, test videos $\mathcal{T}$, grid $\Lambda$
\Ensure scores $\hat{\mathbf{y}}_t$ for the frames of $\mathcal{T}$
\For{each frame $t$ of $\mathcal{F}$, $\mathcal{V}$ and $\mathcal{T}$}
\State $\mathbf{u}_t\gets[\mathbf{z}_t;\mathbf{a}_t]$ \Comment{outputs of $f$}
\State $\bm{\xi}_t\gets[\mathbf{u}_t;\bar{\mathbf{u}}_t^{(1)};\ldots;\bar{\mathbf{u}}_t^{(M)}]$ \Comment{\cref{eq:mean,eq:xi}}
\EndFor
\State standardise every $\bm{\xi}_t$ with the statistics of $\mathcal{F}$
\State $\mathbf{Q}_\mathcal{F},\mathbf{Y}_\mathcal{F}\gets$ rows $\bm{\xi}_t$, $\tilde{\mathbf{y}}_t$ of $\mathcal{F}$
\State $\mathbf{Q}_\mathcal{V},\mathbf{Y}_\mathcal{V}\gets$ rows $\bm{\xi}_t$, $\tilde{\mathbf{y}}_t$ of $\mathcal{V}$
\State $\mathbf{U}\mathbf{S}\mathbf{U}^{\top}\gets\mathrm{eig}(\mathbf{Q}_\mathcal{F}^{\top}\mathbf{Q}_\mathcal{F})$
\For{$\lambda\in\Lambda$}
\State $\mathbf{W}_\lambda\gets \mathbf{U}(\mathbf{S}+\lambda \mathbf{I})^{-1}\mathbf{U}^{\top}\mathbf{Q}_\mathcal{F}^{\top}\mathbf{Y}_\mathcal{F}$ \Comment{\cref{eq:cf}}
\EndFor
\State $\lambda^*\gets\arg\max_\lambda\mathrm{AP}_\mathrm{IVT}(\mathbf{Y}_\mathcal{V},\mathbf{Q}_\mathcal{V}\mathbf{W}_\lambda)$
\State $\mathbf{W}\gets\mathbf{W}_{\lambda^*}$
\For{each frame $t$ of $\mathcal{T}$}
\State $\hat{\mathbf{y}}_t\gets\mathbf{W}^{\top}\bm{\xi}_t$ \Comment{\cref{eq:inference}}
\State $\hat y^\mathrm{IVT}_{t,c}\gets$ \cref{eq:supp-zero} for every triplet $c$ absent from $\mathcal{F}$
\EndFor
\end{algorithmic}
\end{algorithm}

\paragraph{Absent-triplet rule.} A triplet without positive fitting frames has a zero target column, so the readout assigns it a constant score. We replace this score by the product of the predicted component scores,
\begin{equation}
    \hat y^\mathrm{IVT}_{t,c}=q\big(\hat y^\mathrm{I}_{t,i(c)}\big)\,q\big(\hat y^\mathrm{V}_{t,v(c)}\big)\,q\big(\hat y^\mathrm{T}_{t,\tau(c)}\big),
    \label{eq:supp-zero}
\end{equation}
where $q(a)=\min\big(\max(a,10^{-4}),1\big)$ clips each component score to $[10^{-4},1]$. The rule is part of STRIDE and only affects the triplets that never occur in the fitting videos. On all other triplets, where it is not applied, STRIDE improves \apivt{} over $f$ by 3.3, 2.6 and 4.1 points on CholecT45, CholecT50 and ProstaTD, averaged over the seven online recognisers of \cref{tab:table1}.

\section{Out-of-fold fitting}
\label{sec:supp-crossfit}
STRIDE fits $\mathbf{W}$ on the fitting frames on which $f$ itself was trained~(\cref{sec:stride}). Stacked learning instead fits the second stage on out-of-fold predictions~\cite{wolpert1992stacked,cohen2005stacked}, and we compare the two on CholecT50. Within each fold, we split the fitting videos into five groups and train five further models of $f$, each on four of the groups. Every fitting video then has outputs from a model that was not trained on it. We then fit STRIDE with the out-of-fold logits $\mathbf{a}_t$ of the fitting videos. Validation and test outputs always come from $f$ itself, and $\lambda$ is selected on the validation videos as before.

\Cref{tab:supp_crossfit} reports the change in \apivt{} against STRIDE fitted on the outputs of $f$ itself. Out-of-fold fitting improves none of the four recognisers beyond the noise floor, as \apivt{} changes by between $-$2.0 and $+$0.2 points. We therefore fit STRIDE on the outputs of $f$ itself in all experiments.

\begin{table}[htbp]\centering\footnotesize\setlength{\tabcolsep}{8pt}
\begin{tabular}{lc}
\toprule
Recogniser & $\Delta$\,\apivt{} \\
\midrule
SurgeNetXL~\cite{jaspers2025surgenetxl} & $+$0.2 \\
EndoViT~\cite{batic2024endovit} & $-$0.2 \\
TrackMAE~\cite{vandeghen2026trackmae} & $-$0.3 \\
SPIRIT-S~\cite{sharma2026spirit} & $-$2.0 \\
\bottomrule
\end{tabular}
\caption{Change in \apivt{} on CholecT50 when STRIDE is fitted on out-of-fold logits instead of the logits of $f$ itself. SurgeNetXL and TrackMAE are the adapted encoders of \cref{tab:table2}, and EndoViT and SPIRIT-S are the recognisers of \cref{tab:table1}. TrackMAE is evaluated offline and the others online.}
\label{tab:supp_crossfit}
\end{table}

\section{Protocol and implementation}
\label{sec:supp-recipes}
\paragraph{Evaluation.} We compute all metrics with \texttt{ivtmetrics}~\cite{nwoye2022splits}. The AP of each class is computed within each test video, averaged over the videos in which the class occurs, and then averaged over classes. All triplet classes are evaluated, including those with a null verb or target. AP$_\mathrm{I}$, AP$_\mathrm{V}$ and AP$_\mathrm{T}$ follow the convention of \texttt{ivtmetrics} for $f$ and STRIDE alike.

\paragraph{Splits.} All three datasets use their official five folds. On CholecT45 and CholecT50, we take five validation videos per fold evenly spaced in the list of training videos, which leaves 31 and 35 fitting videos. ProstaTD uses the two official validation videos of each fold, which leaves 14 or 15 fitting videos. Within each fold, $\lambda$ and the training epoch of every existing recogniser we retrain are selected on the validation videos. The temporal scales are fixed~(\cref{sec:supp-window}).

\paragraph{STRIDE.} A recogniser that outputs probabilities $\mathbf{p}_t$ enters through the logits $\mathbf{a}_t=\log\big(\mathbf{p}_t/(1-\mathbf{p}_t)\big)$. The readout has $K=131$ targets on CholecT45 and CholecT50, and $K=116$ on ProstaTD. A fit takes about 1.5\,min per fold on four CPU threads for TrackMAE, whose features have 2048 dimensions. Across the thirteen encoders on CholecT45 and ProstaTD, the complete procedure for one fold, including loading the features and scoring, takes between two and six minutes on a CPU.

\paragraph{Recognisers.} On the CholecT45 and CholecT50 datasets, Tripnet, Attention-Tripnet and Rendezvous use the checkpoints released for the official folds and are retrained where no checkpoint is released. SDSwin uses its released checkpoints. CurConMix+ uses its released encoders, and we retrain its temporal module. EndoViT is fine-tuned from its released pre-trained weights. TERL and all recognisers on the ProstaTD dataset are trained by us. 

\paragraph{SPIRIT.} We re-implement SPIRIT~\cite{sharma2026spirit} from its description, as it has no public implementation. The encoder is DINOv3 ViT-L/16~\cite{simeoni2025dinov3} at 224\,px, with its last eight blocks trainable. Each clip holds eight frames at 1\,fps and ends at the current frame. The $14\times14$ spatial tokens of each frame are average-pooled to $8\times8$, which feed the unary, pairwise and triplet heads of the original. We train for 30 epochs with AdamW and select the epoch on the validation videos. SPIRIT-TS adds the distillation stage, in which a frozen SPIRIT-S teacher supervises a re-initialised student.

\paragraph{Adaptation.} Following the linear evaluation protocol of DINOv2~\cite{oquab2023dinov2}, the feature $\mathbf{z}_t$ of an image encoder concatenates the $\ell_2$-normalised class token with the mean of the patch tokens. Frames are resized and centre-cropped to 336\,px. SigLIP2 has no class token and uses its pooled embedding in its place. LemonFM and PeskaVLP are convolutional networks, so their feature is the globally pooled vector of the encoder. LoRA is applied to the pointwise linear layers of LemonFM and to the $1\times1$ convolutions of PeskaVLP. The feature of a video encoder concatenates the mean token of the central time step of the clip with the mean of all tokens. LoRA and the linear classifier are trained per fold on the fitting videos only. The binary cross-entropy weights the positive term of class $k$ by $(1-\rho_k)/\rho_k$, clipped to $[1,50]$, where $\rho_k$ is the frequency of the class. We use a one-cycle schedule with 10\% warm-up, a weight decay of $10^{-4}$, and random resized crops, flips and colour jitter.

\paragraph{Trained readouts.} The trained readouts of \cref{tab:abl_readout} use the same fitting and validation videos as STRIDE. MS-TCN++~\cite{li2020mstcnpp} has four stages of ten dilated residual layers with 64 channels and causal convolutions. It is trained with binary cross-entropy and the temporal smoothing loss, using Adam at a learning rate of $5\cdot10^{-4}$ for up to 300 epochs, with early stopping on the validation videos. The MLP has one hidden layer of 1024 units and is trained for 30 epochs with AdamW, with the learning rate and weight decay selected on the validation videos. Both add their output to the logits of $f$ and start from the predictions of $f$. Logistic regression is fitted to convergence, with its L2 penalty selected on the validation videos in the same way as $\lambda$.

\section{Statistics}
\label{sec:supp-stats}
The main paper calls an improvement reliable if it exceeds a noise floor of 1.62 \apivt{} and is positive on at least four folds~(\cref{sec:impl}).

\paragraph{Noise floor.} The noise floor is the standard deviation of the difference in five-fold \apivt{} between two runs that differ only in their random seed. A change below it could come from the seed alone. We measured a standard deviation of at most 1.62 for this difference and use it as the floor.

\paragraph{Intervals.} Comparisons are paired within test folds. For each recogniser and dataset, the gain of STRIDE is computed per fold. \Cref{tab:supp_ci_table1,tab:supp_ci_table2} report its mean over the five folds and a 95\% $t$-interval over the folds (four degrees of freedom) for the two main results. In \cref{tab:supp_ci_table1}, all 22 gains beyond the noise floor are positive on all five folds, and their intervals exclude zero.

The two criteria agree for 35 of the 39 pairs in \cref{tab:supp_ci_table2}, and the other four are on ProstaTD. For LemonFM, SigLIP2-L and SurgRec-JEPA there, the interval excludes zero. By the interval alone, these three gains would count as improvements. Their gains are below the noise floor, however, so we report them as within noise. The noise floor is therefore the stricter criterion in three of the four cases. For DINOv3, the gain exceeds the noise floor but varies across folds, and its interval includes zero. The gains on ProstaTD are smaller than on the Cholec datasets, so they lie closer to both criteria.

\section{Label statistics}
\label{sec:supp-labels}
The main paper relates the small change of the offline recognisers on ProstaTD to its shorter interactions~(\cref{sec:exp_methods}). \Cref{tab:label_stats} reports the label statistics behind this statement. Labelled segments are shorter on ProstaTD, with a median of 4\,s against 10\,s on CholecT45 and 9\,s on CholecT50. ProstaTD also has more triplets active at the same time, 2.7 per frame against 1.5 on the Cholec datasets.

\providecolor{dgrey}{gray}{0.6}\providecolor{dup}{HTML}{2E8B57}
\begin{table*}[t]\centering\footnotesize\setlength{\tabcolsep}{8pt}
\begin{tabular}{l@{\hspace{24pt}}rc@{\hspace{24pt}}rc@{\hspace{24pt}}rc}
\toprule
 & \multicolumn{2}{c}{CholecT45} & \multicolumn{2}{c}{CholecT50} & \multicolumn{2}{c}{ProstaTD} \\
\cmidrule(lr){2-3}\cmidrule(lr){4-5}\cmidrule(lr){6-7}
Recogniser & $\delta$ & 95\% interval & $\delta$ & 95\% interval & $\delta$ & 95\% interval \\
\midrule
{\textbf{Online recognisers}} \\ \midrule
Tripnet~\cite{nwoye2020tripnet} & \textcolor{dup}{\textbf{+7.37}} & [+4.13, +10.61]  & \textcolor{dup}{\textbf{+5.30}} & [+3.26, +7.33]  & \textcolor{dup}{\textbf{+6.88}} & [+5.28, +8.47]  \\
Attention-Tripnet~\cite{nwoye2022rendezvous} & \textcolor{dup}{\textbf{+5.85}} & [+3.52, +8.18]  & \textcolor{dup}{\textbf{+4.51}} & [+3.82, +5.21]  & \textcolor{dup}{\textbf{+6.67}} & [+5.79, +7.54]  \\
Rendezvous~\cite{nwoye2022rendezvous} & \textcolor{dup}{\textbf{+2.64}} & [+1.67, +3.61]  & \textcolor{dup}{\textbf{+3.10}} & [+1.43, +4.78]  & \textcolor{dup}{\textbf{+5.95}} & [+5.30, +6.61]  \\
SDSwin~\cite{yamlahi2023sdswin} & \textcolor{dup}{\textbf{+3.41}} & [+1.36, +5.46]  & \textcolor{dup}{\textbf{+2.70}} & [+0.78, +4.61]  & \textcolor{dup}{\textbf{+2.15}} & [+0.99, +3.30]  \\
EndoViT~\cite{batic2024endovit} & \textcolor{dup}{\textbf{+5.49}} & [+3.81, +7.17]  & \textcolor{dup}{\textbf{+3.86}} & [+3.00, +4.71]  & \textcolor{dup}{\textbf{+2.62}} & [+1.63, +3.61]  \\
SPIRIT-S~\cite{sharma2026spirit} & \textcolor{dup}{\textbf{+1.66}} & [+0.38, +2.94]  & \textcolor{dup}{\textbf{+1.71}} & [+0.39, +3.03]  & \textcolor{dup}{\textbf{+2.42}} & [+1.09, +3.74]  \\
SPIRIT-TS~\cite{sharma2026spirit} & \textcolor{dgrey}{\textbf{$-$0.11}} & [$-$1.54, +1.32]  & \textcolor{dgrey}{\textbf{$-$0.10}} & [$-$2.95, +2.75]  & \textcolor{dgrey}{\textbf{+0.50}} & [$-$0.88, +1.87]  \\
\midrule
{\textbf{Offline recognisers}} \\ \midrule
TERL~\cite{gui2024terl} & \textcolor{dup}{\textbf{+4.79}} & [+2.45, +7.12]  & \textcolor{dup}{\textbf{+3.13}} & [+1.80, +4.46]  & \textcolor{dgrey}{\textbf{$-$0.64}} & [$-$1.69, +0.41]  \\
CurConMix+~\cite{jeon2026curconmixplus} & \textcolor{dup}{\textbf{+3.81}} & [+2.32, +5.29]  & \textcolor{dup}{\textbf{+3.48}} & [+2.43, +4.53]  & \textcolor{dgrey}{\textbf{+0.78}} & [0.00, +1.57]  \\
\bottomrule
\end{tabular}
\caption{Paired 95\% confidence intervals for \cref{tab:table1}. $\delta$ is the change in \apivt{} after adding STRIDE, averaged over the five official cross-validation folds, with its $t$-interval over the folds. Changes below the noise floor ($|\delta| < 1.62$) are shown in grey.}
\label{tab:supp_ci_table1}
\end{table*}

\begin{table*}[t]\centering\footnotesize\setlength{\tabcolsep}{8pt}
\begin{tabular}{l@{\hspace{24pt}}rc@{\hspace{24pt}}rc@{\hspace{24pt}}rc}
\toprule
 & \multicolumn{2}{c}{CholecT45} & \multicolumn{2}{c}{CholecT50} & \multicolumn{2}{c}{ProstaTD} \\
\cmidrule(lr){2-3}\cmidrule(lr){4-5}\cmidrule(lr){6-7}
Encoder & $\delta$ & 95\% interval & $\delta$ & 95\% interval & $\delta$ & 95\% interval \\
\midrule
{\textbf{Online image encoders}} \\ \midrule
EndoViT~\cite{batic2024endovit} & \textcolor{dup}{\textbf{+5.37}} & [+2.51, +8.24] & \textcolor{dup}{\textbf{+4.34}} & [+2.60, +6.08] & \textcolor{dup}{\textbf{+3.51}} & [+2.92, +4.10] \\
PeskaVLP~\cite{yuan2024peskavlp} & \textcolor{dup}{\textbf{+7.16}} & [+4.53, +9.79] & \textcolor{dup}{\textbf{+7.07}} & [+4.82, +9.33] & \textcolor{dup}{\textbf{+2.87}} & [+1.24, +4.51] \\
LemonFM~\cite{che2026lemon} & \textcolor{dup}{\textbf{+3.19}} & [+1.88, +4.50] & \textcolor{dup}{\textbf{+1.63}} & [+0.06, +3.21] & \textcolor{dgrey}{\textbf{+1.09}} & [+0.50, +1.68] \\
SigLIP2-L~\cite{tschannen2025siglip2} & \textcolor{dup}{\textbf{+5.54}} & [+4.21, +6.88] & \textcolor{dup}{\textbf{+4.27}} & [+3.40, +5.14] & \textcolor{dgrey}{\textbf{+1.37}} & [+0.07, +2.68] \\
DINOv3~\cite{simeoni2025dinov3} & \textcolor{dup}{\textbf{+3.84}} & [+2.52, +5.16] & \textcolor{dup}{\textbf{+3.73}} & [+1.77, +5.69] & \textcolor{dup}{\textbf{+1.70}} & [$-$0.18, +3.57] \\
SurgLIME~\cite{che2026surglime} & \textcolor{dup}{\textbf{+4.31}} & [+2.70, +5.92] & \textcolor{dup}{\textbf{+4.18}} & [+2.24, +6.13] & \textcolor{dup}{\textbf{+2.42}} & [+0.46, +4.39] \\
SurgeNetXL~\cite{jaspers2025surgenetxl} & \textcolor{dup}{\textbf{+3.88}} & [+2.10, +5.66] & \textcolor{dup}{\textbf{+3.27}} & [+1.79, +4.75] & \textcolor{dup}{\textbf{+3.24}} & [+1.51, +4.98] \\
PL-Stitch~\cite{che2026plstitch} & \textcolor{dup}{\textbf{+4.46}} & [+3.65, +5.27] & \textcolor{dup}{\textbf{+4.44}} & [+2.17, +6.72] & \textcolor{dup}{\textbf{+2.45}} & [+0.94, +3.95] \\
\midrule
{\textbf{Offline video encoders}} \\ \midrule
VideoMAE V2~\cite{wang2023videomaev2} & \textcolor{dup}{\textbf{+2.34}} & [+1.23, +3.44] & \textcolor{dgrey}{\textbf{+1.11}} & [$-$1.17, +3.39] & \textcolor{dgrey}{\textbf{+0.66}} & [$-$0.63, +1.95] \\
V-JEPA 2~\cite{assran2025vjepa2} & \textcolor{dup}{\textbf{+5.05}} & [+2.74, +7.36] & \textcolor{dup}{\textbf{+3.95}} & [+2.85, +5.05] & \textcolor{dgrey}{\textbf{+0.52}} & [$-$1.07, +2.11] \\
InternVideo2~\cite{wang2024internvideo2} & \textcolor{dup}{\textbf{+2.60}} & [+1.05, +4.15] & \textcolor{dup}{\textbf{+2.14}} & [+0.27, +4.00] & \textcolor{dgrey}{\textbf{$-$0.62}} & [$-$1.49, +0.26] \\
TrackMAE~\cite{vandeghen2026trackmae} & \textcolor{dup}{\textbf{+3.17}} & [+1.43, +4.90] & \textcolor{dup}{\textbf{+2.35}} & [+1.15, +3.56] & \textcolor{dgrey}{\textbf{$-$0.17}} & [$-$1.80, +1.45] \\
SurgRec-JEPA$^\dagger$~\cite{lu2026surgrec} & \textcolor{dup}{\textbf{+4.52}} & [+2.63, +6.41] & \textcolor{dup}{\textbf{+3.71}} & [+1.86, +5.56] & \textcolor{dgrey}{\textbf{+1.36}} & [+0.15, +2.57] \\
\bottomrule
\end{tabular}
\caption{Paired 95\% confidence intervals for \cref{tab:table2}. $\delta$ is the change in \apivt{} after adding STRIDE, averaged over the five official cross-validation folds, with its $t$-interval over the folds. Changes below the noise floor ($|\delta| < 1.62$) are shown in grey. $^\dagger\!$ Pre-training contains all CholecT50 test videos.}
\label{tab:supp_ci_table2}
\end{table*}

\begin{table}[htbp]\centering\footnotesize\setlength{\tabcolsep}{6pt}
\begin{tabular}{lccc}
\toprule
 & CholecT45 & CholecT50 & ProstaTD \\
\midrule
Median segment length (s) & 10 & 9 & 4 \\
\midrule
Triplets per frame & 1.5 & 1.5 & 2.7 \\
Frames with a triplet (\%) & 91 & 89 & 99.6 \\
\bottomrule
\end{tabular}
\caption{Label statistics of the three datasets at 1\,fps, over all annotated videos. A labelled segment is a run of consecutive frames in which one triplet class is active.}
\label{tab:label_stats}
\end{table}

\section{Foundation encoders in the offline setting}
\label{sec:supp-settings} \Cref{sec:exp_encoders} states that the smaller gain of video encoders is not due to the offline setting. \Cref{tab:supp_settings} supports this by evaluating all thirteen encoders offline. In the offline setting, STRIDE improves \apivt{} for all eight image encoders on every dataset. Image encoders gain 6.5, 5.6 and 2.5 points on average on CholecT45, CholecT50 and ProstaTD, against 3.3, 2.4 and 0.1 points for video encoders without SurgRec-JEPA. Image encoders therefore still gain more when both groups are evaluated offline, which is consistent with STRIDE helping most where the encoder has the least temporal context.

\begin{table*}[t]\centering\footnotesize\setlength{\tabcolsep}{4pt}
\begin{tabular}{l@{\hspace{24pt}}rrr>{\columncolor{ivtbase}}r@{\hspace{24pt}}rrr>{\columncolor{ivtbase}}r@{\hspace{24pt}}rrr>{\columncolor{ivtbase}}r}
\toprule
 & \multicolumn{4}{c}{CholecT45} & \multicolumn{4}{c}{CholecT50} & \multicolumn{4}{c}{ProstaTD} \\
\cmidrule(lr){2-5}\cmidrule(lr){6-9}\cmidrule(lr){10-13}
Encoder & AP$_\mathrm{I}$ & AP$_\mathrm{V}$ & AP$_\mathrm{T}$ & AP$_\mathrm{IVT}$ & AP$_\mathrm{I}$ & AP$_\mathrm{V}$ & AP$_\mathrm{T}$ & AP$_\mathrm{IVT}$ & AP$_\mathrm{I}$ & AP$_\mathrm{V}$ & AP$_\mathrm{T}$ & AP$_\mathrm{IVT}$ \\
\midrule
{\textbf{Image encoders}} \\ \midrule
EndoViT~\cite{batic2024endovit} & 77.3 & 56.0 & 43.8 & 28.0\phantom{$^{\text{\scriptsize +0.0}}$} & 79.1 & 56.7 & 43.3 & 28.0\phantom{$^{\text{\scriptsize +0.0}}$} & 85.0 & 60.2 & 55.7 & 28.6\phantom{$^{\text{\scriptsize +0.0}}$} \\
\quad + STRIDE & 86.3 & 65.6 & 49.7 & \cellcolor{ivtstride}35.2$^{\textcolor{dup}{\text{\scriptsize \textbf{+7.2}}}}$ & 87.9 & 65.2 & 49.0 & \cellcolor{ivtstride}34.4$^{\textcolor{dup}{\text{\scriptsize \textbf{+6.3}}}}$ & 86.7 & 65.1 & 60.3 & \cellcolor{ivtstride}31.4$^{\textcolor{dup}{\text{\scriptsize \textbf{+2.8}}}}$ \\
\midrule
PeskaVLP~\cite{yuan2024peskavlp} & 78.6 & 58.4 & 41.0 & 26.2\phantom{$^{\text{\scriptsize +0.0}}$} & 78.8 & 57.6 & 40.9 & 25.4\phantom{$^{\text{\scriptsize +0.0}}$} & 82.9 & 59.4 & 55.1 & 28.5\phantom{$^{\text{\scriptsize +0.0}}$} \\
\quad + STRIDE & 90.1 & 68.0 & 47.5 & \cellcolor{ivtstride}34.7$^{\textcolor{dup}{\text{\scriptsize \textbf{+8.6}}}}$ & 89.9 & 66.2 & 48.0 & \cellcolor{ivtstride}33.8$^{\textcolor{dup}{\text{\scriptsize \textbf{+8.4}}}}$ & 87.7 & 65.1 & 62.2 & \cellcolor{ivtstride}32.2$^{\textcolor{dup}{\text{\scriptsize \textbf{+3.7}}}}$ \\
\midrule
LemonFM~\cite{che2026lemon} & 87.4 & 66.6 & 48.4 & 34.1\phantom{$^{\text{\scriptsize +0.0}}$} & 87.5 & 66.1 & 48.0 & 33.9\phantom{$^{\text{\scriptsize +0.0}}$} & 91.7 & 68.9 & 64.9 & 37.1\phantom{$^{\text{\scriptsize +0.0}}$} \\
\quad + STRIDE & 92.7 & 72.0 & 52.2 & \cellcolor{ivtstride}39.6$^{\textcolor{dup}{\text{\scriptsize \textbf{+5.4}}}}$ & 92.6 & 71.0 & 50.9 & \cellcolor{ivtstride}36.9$^{\textcolor{dup}{\text{\scriptsize \textbf{+3.0}}}}$ & 92.1 & 71.4 & 67.7 & \cellcolor{ivtstride}38.3$^{\textcolor{dgrey}{\text{\scriptsize \textbf{+1.2}}}}$ \\
\midrule
SigLIP2-L~\cite{tschannen2025siglip2} & 88.7 & 66.2 & 46.8 & 33.4\phantom{$^{\text{\scriptsize +0.0}}$} & 88.6 & 65.5 & 46.8 & 31.9\phantom{$^{\text{\scriptsize +0.0}}$} & 91.2 & 68.5 & 66.2 & 37.6\phantom{$^{\text{\scriptsize +0.0}}$} \\
\quad + STRIDE & 93.6 & 71.2 & 51.4 & \cellcolor{ivtstride}40.6$^{\textcolor{dup}{\text{\scriptsize \textbf{+7.1}}}}$ & 93.4 & 69.9 & 51.9 & \cellcolor{ivtstride}37.5$^{\textcolor{dup}{\text{\scriptsize \textbf{+5.6}}}}$ & 93.1 & 71.1 & 68.7 & \cellcolor{ivtstride}39.7$^{\textcolor{dup}{\text{\scriptsize \textbf{+2.1}}}}$ \\
\midrule
DINOv3~\cite{simeoni2025dinov3} & 89.4 & 66.9 & 47.4 & 33.5\phantom{$^{\text{\scriptsize +0.0}}$} & 88.9 & 66.0 & 47.8 & 33.0\phantom{$^{\text{\scriptsize +0.0}}$} & 91.1 & 69.0 & 64.5 & 37.7\phantom{$^{\text{\scriptsize +0.0}}$} \\
\quad + STRIDE & 93.1 & 72.2 & 52.2 & \cellcolor{ivtstride}39.3$^{\textcolor{dup}{\text{\scriptsize \textbf{+5.9}}}}$ & 92.7 & 70.2 & 53.1 & \cellcolor{ivtstride}38.0$^{\textcolor{dup}{\text{\scriptsize \textbf{+5.0}}}}$ & 92.3 & 72.1 & 68.7 & \cellcolor{ivtstride}39.7$^{\textcolor{dup}{\text{\scriptsize \textbf{+2.0}}}}$ \\
\midrule
SurgLIME~\cite{che2026surglime} & 87.7 & 66.2 & 48.3 & 34.1\phantom{$^{\text{\scriptsize +0.0}}$} & 87.4 & 65.5 & 47.6 & 33.3\phantom{$^{\text{\scriptsize +0.0}}$} & 91.9 & 70.3 & 66.2 & 38.5\phantom{$^{\text{\scriptsize +0.0}}$} \\
\quad + STRIDE & 93.1 & 72.4 & 51.4 & \cellcolor{ivtstride}40.0$^{\textcolor{dup}{\text{\scriptsize \textbf{+5.9}}}}$ & 92.4 & 70.6 & 51.5 & \cellcolor{ivtstride}38.9$^{\textcolor{dup}{\text{\scriptsize \textbf{+5.6}}}}$ & 92.4 & 72.9 & 69.2 & \cellcolor{ivtstride}40.8$^{\textcolor{dup}{\text{\scriptsize \textbf{+2.3}}}}$ \\
\midrule
SurgeNetXL~\cite{jaspers2025surgenetxl} & 89.2 & 67.2 & 49.6 & 35.1\phantom{$^{\text{\scriptsize +0.0}}$} & 89.1 & 66.1 & 50.3 & 34.4\phantom{$^{\text{\scriptsize +0.0}}$} & 93.5 & 69.2 & 66.0 & 38.0\phantom{$^{\text{\scriptsize +0.0}}$} \\
\quad + STRIDE & 93.1 & 73.4 & 54.0 & \cellcolor{ivtstride}41.0$^{\textcolor{dup}{\text{\scriptsize \textbf{+5.9}}}}$ & 93.4 & 72.2 & 54.6 & \cellcolor{ivtstride}39.4$^{\textcolor{dup}{\text{\scriptsize \textbf{+5.0}}}}$ & 94.4 & 72.8 & 69.9 & \cellcolor{ivtstride}40.9$^{\textcolor{dup}{\text{\scriptsize \textbf{+2.9}}}}$ \\
\midrule
PL-Stitch~\cite{che2026plstitch} & 87.8 & 66.8 & 48.2 & 34.7\phantom{$^{\text{\scriptsize +0.0}}$} & 87.0 & 65.0 & 47.4 & 33.9\phantom{$^{\text{\scriptsize +0.0}}$} & 91.8 & 70.3 & 66.2 & 38.8\phantom{$^{\text{\scriptsize +0.0}}$} \\
\quad + STRIDE & 93.0 & 72.9 & 52.2 & \cellcolor{ivtstride}40.9$^{\textcolor{dup}{\text{\scriptsize \textbf{+6.3}}}}$ & 92.1 & 70.6 & 52.5 & \cellcolor{ivtstride}40.0$^{\textcolor{dup}{\text{\scriptsize \textbf{+6.1}}}}$ & 92.5 & 73.0 & 68.5 & \cellcolor{ivtstride}41.4$^{\textcolor{dup}{\text{\scriptsize \textbf{+2.6}}}}$ \\
\midrule
{\textbf{Video encoders}} \\ \midrule
VideoMAE V2~\cite{wang2023videomaev2} & 91.0 & 69.8 & 50.1 & 38.1\phantom{$^{\text{\scriptsize +0.0}}$} & 91.0 & 68.3 & 50.8 & 37.8\phantom{$^{\text{\scriptsize +0.0}}$} & 88.6 & 74.1 & 66.0 & 37.8\phantom{$^{\text{\scriptsize +0.0}}$} \\
\quad + STRIDE & 90.9 & 70.3 & 51.4 & \cellcolor{ivtstride}40.5$^{\textcolor{dup}{\text{\scriptsize \textbf{+2.3}}}}$ & 91.5 & 69.2 & 51.8 & \cellcolor{ivtstride}38.9$^{\textcolor{dgrey}{\text{\scriptsize \textbf{+1.1}}}}$ & 87.5 & 74.7 & 67.2 & \cellcolor{ivtstride}38.5$^{\textcolor{dgrey}{\text{\scriptsize \textbf{+0.7}}}}$ \\
\midrule
V-JEPA 2~\cite{assran2025vjepa2} & 90.3 & 72.7 & 52.5 & 38.5\phantom{$^{\text{\scriptsize +0.0}}$} & 90.5 & 70.8 & 51.3 & 37.6\phantom{$^{\text{\scriptsize +0.0}}$} & 88.6 & 75.9 & 68.8 & 40.8\phantom{$^{\text{\scriptsize +0.0}}$} \\
\quad + STRIDE & 92.2 & 74.2 & 53.5 & \cellcolor{ivtstride}43.5$^{\textcolor{dup}{\text{\scriptsize \textbf{+5.1}}}}$ & 92.2 & 72.4 & 53.7 & \cellcolor{ivtstride}41.6$^{\textcolor{dup}{\text{\scriptsize \textbf{+3.9}}}}$ & 88.8 & 77.8 & 70.6 & \cellcolor{ivtstride}41.4$^{\textcolor{dgrey}{\text{\scriptsize \textbf{+0.5}}}}$ \\
\midrule
InternVideo2~\cite{wang2024internvideo2} & 91.8 & 72.0 & 51.7 & 40.2\phantom{$^{\text{\scriptsize +0.0}}$} & 92.6 & 70.9 & 51.9 & 39.6\phantom{$^{\text{\scriptsize +0.0}}$} & 90.5 & 74.7 & 70.3 & 40.7\phantom{$^{\text{\scriptsize +0.0}}$} \\
\quad + STRIDE & 93.0 & 72.1 & 52.3 & \cellcolor{ivtstride}42.8$^{\textcolor{dup}{\text{\scriptsize \textbf{+2.6}}}}$ & 93.1 & 71.6 & 52.6 & \cellcolor{ivtstride}41.7$^{\textcolor{dup}{\text{\scriptsize \textbf{+2.1}}}}$ & 90.3 & 75.5 & 70.8 & \cellcolor{ivtstride}40.1$^{\textcolor{dgrey}{\text{\scriptsize \textbf{$-$0.6}}}}$ \\
\midrule
TrackMAE~\cite{vandeghen2026trackmae} & 93.0 & 75.2 & 53.7 & 41.9\phantom{$^{\text{\scriptsize +0.0}}$} & 93.2 & 72.4 & 54.3 & 40.7\phantom{$^{\text{\scriptsize +0.0}}$} & 92.5 & 78.5 & 71.6 & 44.8\phantom{$^{\text{\scriptsize +0.0}}$} \\
\quad + STRIDE & 93.2 & 75.1 & 54.0 & \cellcolor{ivtstride}45.1$^{\textcolor{dup}{\text{\scriptsize \textbf{+3.2}}}}$ & 93.4 & 72.8 & 54.1 & \cellcolor{ivtstride}43.0$^{\textcolor{dup}{\text{\scriptsize \textbf{+2.4}}}}$ & 92.2 & 79.4 & 72.1 & \cellcolor{ivtstride}44.6$^{\textcolor{dgrey}{\text{\scriptsize \textbf{$-$0.2}}}}$ \\
\midrule
SurgRec-JEPA$^\dagger$~\cite{lu2026surgrec} & 93.6 & 78.3 & 58.0 & 46.2\phantom{$^{\text{\scriptsize +0.0}}$} & 93.5 & 75.8 & 57.7 & 44.4\phantom{$^{\text{\scriptsize +0.0}}$} & 92.7 & 80.2 & 73.2 & 47.0\phantom{$^{\text{\scriptsize +0.0}}$} \\
\quad + STRIDE & 94.1 & 79.3 & 58.8 & \cellcolor{ivtstride}50.7$^{\textcolor{dup}{\text{\scriptsize \textbf{+4.5}}}}$ & 94.1 & 77.4 & 59.2 & \cellcolor{ivtstride}48.1$^{\textcolor{dup}{\text{\scriptsize \textbf{+3.7}}}}$ & 93.5 & 81.7 & 74.5 & \cellcolor{ivtstride}48.3$^{\textcolor{dgrey}{\text{\scriptsize \textbf{+1.4}}}}$ \\
\bottomrule
\end{tabular}
\caption{STRIDE on foundation encoders adapted with LoRA and a linear classifier, evaluated in the offline setting. Mean AP (\%) over the official cross-validation folds. Image encoders use the adapters of \cref{tab:table2} with centred windows, and the video-encoder rows are those of \cref{tab:table2}. Superscript numbers: (LoRA + STRIDE) $-$ LoRA, grey for $|\delta|$$\,<\,$$1.62$, i.e. noise floor. $^\dagger\!$ Pre-training contains all CholecT50 test videos.}
\label{tab:supp_settings}
\end{table*}

\section{Temporal scales}
\label{sec:supp-window}
The main paper ablates the temporal scales with four configurations~(\cref{tab:abl_windows}), and \cref{tab:supp_window_all} extends this ablation with further sets of scales. Every set improves on $f$. On CholecT50, a single scale performs best when it is short, with 34.6 \apivt{} at 2\,s against 33.5 at 128\,s. Combining the 2\,s and 32\,s scales improves on either scale alone on both datasets. The sets without the 2\,s scale are lower on CholecT50. A fourth scale changes \apivt{} by at most 0.2 points.

\begin{table}[htbp]\centering\footnotesize\setlength{\tabcolsep}{12pt}
\begin{tabular}{cl>{\columncolor{ivtbase}}c>{\columncolor{ivtbase}}c}
\toprule
$M$ & $\{\omega_j\}$ (s) & CholecT50 & ProstaTD \\
\midrule
1 & 2 & 34.6 & 34.2 \\
1 & 8 & 34.4 & 34.6 \\
1 & 32 & 33.9 & 34.5 \\
1 & 128 & 33.5 & 34.0 \\
\midrule
2 & 2, 8 & 35.0 & 34.8 \\
2 & 2, 32 & 35.1 & 35.0 \\
2 & 8, 32 & 34.6 & 35.0 \\
\midrule
3 & 2, 4, 8 & 34.8 & 34.7 \\
3 & 2, 8, 32 & \cellcolor{ivtstride}35.2 & \cellcolor{ivtstride}35.2 \\
3 & 8, 32, 128 & 34.5 & 35.1 \\
\midrule
4 & 2, 8, 32, 128 & 35.0 & 35.3 \\
\midrule
$f$ & (baseline) & 32.2 & 31.3 \\
\bottomrule
\end{tabular}
\caption{Further sets of temporal scales, evaluated using \apivt{} and averaged over the online recognisers. The scales used in the paper are highlighted.}
\label{tab:supp_window_all}
\end{table}

\section{Smoothing baselines}
\label{sec:supp-smoothing}
The main paper compares STRIDE with temporal smoothing of the scores of $f$~(\cref{sec:temporal_analysis}). \Cref{tab:abl_smoothing} reports this comparison for all four metrics, averaged over the seven online recognisers of \cref{tab:table1}. The baselines are a causal box filter and an exponential moving average (EMA). The width of the box filter and the decay of the EMA are tuned on the validation videos for every recogniser. STRIDE gives the largest gain in every metric on both datasets.

\begin{table}[htbp]\centering\footnotesize\setlength{\tabcolsep}{2pt}
\resizebox{\linewidth}{!}{%
\begin{tabular}{lccc>{\columncolor{ivtbase}}c@{\hspace{5pt}}ccc>{\columncolor{ivtbase}}c}
\toprule
 & \multicolumn{4}{c}{CholecT50} & \multicolumn{4}{c}{ProstaTD} \\
\cmidrule(lr){2-5}\cmidrule(lr){6-9}
Method & AP$_\mathrm{I}$ & AP$_\mathrm{V}$ & AP$_\mathrm{T}$ & AP$_\mathrm{IVT}$ & AP$_\mathrm{I}$ & AP$_\mathrm{V}$ & AP$_\mathrm{T}$ & AP$_\mathrm{IVT}$ \\
\midrule
$f$ + box filter & 89.4 & 63.2 & 44.3 & 33.1$^{\textcolor{dgrey}{\text{\scriptsize \textbf{+0.9}}}}$ & 88.0 & 64.3 & 58.3 & 31.5$^{\textcolor{dgrey}{\text{\scriptsize \textbf{+0.2}}}}$ \\
$f$ + EMA & 89.9 & 64.2 & 45.0 & 34.0$^{\textcolor{dup}{\text{\scriptsize \textbf{+1.9}}}}$ & 89.2 & 65.4 & 58.9 & 32.8$^{\textcolor{dgrey}{\text{\scriptsize \textbf{+1.5}}}}$ \\
\midrule
STRIDE & 91.1 & 66.4 & 47.6 & \cellcolor{ivtstride}35.2$^{\textcolor{dup}{\text{\scriptsize \textbf{+3.0}}}}$ & 90.1 & 67.7 & 61.6 & \cellcolor{ivtstride}35.2$^{\textcolor{dup}{\text{\scriptsize \textbf{+3.9}}}}$ \\
\midrule
$f$ (baseline) & 88.8 & 63.3 & 44.0 & 32.2\phantom{$^{\text{\scriptsize +0.0}}$} & 88.0 & 64.2 & 57.2 & 31.3\phantom{$^{\text{\scriptsize +0.0}}$} \\
\bottomrule
\end{tabular}}
\caption{Smoothing the scores of $f$ against STRIDE, averaged over the online recognisers. The box filter and EMA are causal. Superscripts on AP$_\mathrm{IVT}$ show the change $\delta$ against $f$, with changes below the noise floor ($|\delta| < 1.62$) shown in grey.}
\label{tab:abl_smoothing}
\end{table}

\section{Class frequency and duration}
\label{sec:supp-freq}
The main paper groups the triplet classes by frequency and by duration~(\cref{sec:freq}). \Cref{tab:supp_freq_dur} extends \cref{tab:abl_freq} to the instrument, verb and target classes, which are split into thirds in the same way.

\paragraph{Frequency.}
STRIDE improves the components most for rare classes. Rare instruments, verbs and targets gain 5.0, 5.4 and 6.5 points, against 1.2 and 2.3 points for frequent verbs and targets. For triplets, the gain is similar in the three groups. AP$_\mathrm{I}$ of the most frequent instruments is the only entry that decreases, from 96.0 to 95.5. These instruments are already recognised almost perfectly, so temporal context has little to add. The window means also respond later at the start and end of an interaction~(\cref{sec:temporal_analysis}), which plausibly accounts for the small decrease. \Cref{tab:table1} shows the same pattern, as AP$_\mathrm{I}$ decreases only for recognisers whose AP$_\mathrm{I}$ already exceeds 93.

\paragraph{Duration.}
Every component improves in every duration group. As for the triplets, the shortest third gains least, 1.8, 2.8 and 3.4 points for instruments, verbs and targets.

\providecommand{\hl}[1]{{\setlength{\fboxsep}{1.0pt}\smash{\colorbox{blue!10}{\vphantom{0}#1}}}}
\begin{table}[htbp]\centering\footnotesize\setlength{\tabcolsep}{2pt}
\begin{tabular}{lcccc}
\toprule
Metric & Lower third & Middle third & Upper third & All \\
\midrule
{\textbf{By frequency}} \\ \midrule
AP$_\mathrm{I}$ & 85.1$\,\to\,$\hl{90.1} & 84.3$\,\to\,$\hl{87.0} & 96.0$\,\to\,$\hl{95.5} & 88.4$\,\to\,$\hl{90.9} \\
AP$_\mathrm{V}$ & 55.2$\,\to\,$\hl{60.6} & 62.8$\,\to\,$\hl{65.6} & 77.0$\,\to\,$\hl{78.2} & 64.0$\,\to\,$\hl{67.4} \\
AP$_\mathrm{T}$ & 31.2$\,\to\,$\hl{37.7} & 52.0$\,\to\,$\hl{54.8} & 63.2$\,\to\,$\hl{65.5} & 48.1$\,\to\,$\hl{52.1} \\
AP$_\mathrm{IVT}$ & 20.3$\,\to\,$\hl{23.7} & 28.4$\,\to\,$\hl{31.8} & 50.7$\,\to\,$\hl{53.9} & 31.9$\,\to\,$\hl{35.5} \\
\midrule
{\textbf{By duration}} \\ \midrule
AP$_\mathrm{I}$ & 87.9$\,\to\,$\hl{89.7} & 85.5$\,\to\,$\hl{88.9} & 92.2$\,\to\,$\hl{94.7} & 88.4$\,\to\,$\hl{90.9} \\
AP$_\mathrm{V}$ & 49.7$\,\to\,$\hl{52.5} & 71.7$\,\to\,$\hl{75.6} & 76.7$\,\to\,$\hl{80.4} & 64.0$\,\to\,$\hl{67.4} \\
AP$_\mathrm{T}$ & 41.8$\,\to\,$\hl{45.2} & 57.1$\,\to\,$\hl{61.4} & 51.1$\,\to\,$\hl{54.9} & 48.1$\,\to\,$\hl{52.1} \\
AP$_\mathrm{IVT}$ & 24.2$\,\to\,$\hl{26.9} & 36.1$\,\to\,$\hl{39.8} & 37.6$\,\to\,$\hl{42.0} & 31.9$\,\to\,$\hl{35.5} \\
\bottomrule
\end{tabular}
\caption{AP before $\to$ after STRIDE for the components and the triplets, averaged over the online recognisers and three datasets. Classes of each kind are split into thirds in each fold, by training frames (rare to frequent) and by segment length (short to long). The column All also includes the classes that never occur in the fitting videos.}
\label{tab:supp_freq_dur}
\end{table}

\begin{table*}[htbp]\centering\footnotesize\setlength{\tabcolsep}{12pt}
\begin{tabular}{lllll>{\columncolor{ivtbase}}c}
\toprule
Method & Venue & Setting & Split & AP computation & AP$_\mathrm{IVT}$ \\
\midrule
{\textbf{CholecT45}} \\ \midrule
RiT~\cite{sharma2023rit} & IJCARS 2023 & online & official folds & unknown & 29.7 \\
CurConMix~\cite{jeon2025curconmix} & MICCAI 2025 & online & official folds & unknown & 39.1 \\
SPIRIT-S~\cite{sharma2026spirit} & arXiv 2026 & online & official folds & unknown & 40.1 \\
TRCoRSurg~\cite{li2026trcorsurg} & CVPR 2026 & online & official folds & unknown & 40.9 \\
SPIRIT-TS~\cite{sharma2026spirit} & arXiv 2026 & online & official folds & unknown & 43.7 \\
MT4MTL-KD~\cite{gui2024mt4mtlkd} & TMI 2024 & offline & official folds & unknown & 37.1 \\
TERL~\cite{gui2024terl} & MICCAI 2024 & offline & official folds & null triplets excluded & 38.9 \\
MEJO~\cite{mejo2025} & arXiv 2025 & offline & official folds & unknown & 41.2 \\
MoeCo~\cite{moeco2026} & TMI 2026 & offline & official folds & unknown & 41.7 \\
CurConMix+~\cite{jeon2026curconmixplus} & TMI 2026 & offline & official folds & null triplets excluded & 42.2 \\
Tripnet~\cite{nwoye2020tripnet} & MICCAI 2020 & online & official folds & \texttt{ivtmetrics} & 24.4 \\
Attention-Tripnet~\cite{nwoye2022rendezvous} & MedIA 2022 & online & official folds & \texttt{ivtmetrics} & 27.2 \\
Rendezvous~\cite{nwoye2022rendezvous} & MedIA 2022 & online & official folds & \texttt{ivtmetrics} & 29.4 \\
SDSwin~\cite{yamlahi2023sdswin} & MICCAI 2023 & online & official folds & \texttt{ivtmetrics} & 36.1 \\
DiffTriplet~\cite{liu2024difftriplet} & arXiv 2024 & online & official folds & \texttt{ivtmetrics} & 40.2 \\
\rowcolor{gray!15}Ours & -- & online & official folds & \texttt{ivtmetrics} & \cellcolor{ivtstride}39.1 \\
\rowcolor{gray!15}Ours & -- & offline & official folds & \texttt{ivtmetrics} & \cellcolor{ivtstride}45.1 \\
\midrule
{\textbf{CholecT50}} \\ \midrule
ForestGCN~\cite{xi2022forestgcn} & TCSVT 2022 & online & Rendezvous split & unknown & 36.7 \\
Chain-of-Look~\cite{xi2023chainoflook} & ACM MM 2023 & online & Rendezvous split & unknown & 38.2 \\
MoeCo~\cite{moeco2026} & TMI 2026 & offline & Rendezvous split & unknown & 40.5 \\
Tripnet~\cite{nwoye2020tripnet} & MICCAI 2020 & online & official folds & \texttt{ivtmetrics} & 25.3 \\
Attention-Tripnet~\cite{nwoye2022rendezvous} & MedIA 2022 & online & official folds & \texttt{ivtmetrics} & 27.2 \\
Rendezvous~\cite{nwoye2022rendezvous} & MedIA 2022 & online & official folds & \texttt{ivtmetrics} & 29.4 \\
DiffTriplet~\cite{liu2024difftriplet} & arXiv 2024 & online & official folds & \texttt{ivtmetrics} & 40.3 \\
\rowcolor{gray!15}Ours & -- & online & official folds & \texttt{ivtmetrics} & \cellcolor{ivtstride}38.4 \\
\rowcolor{gray!15}Ours & -- & offline & official folds & \texttt{ivtmetrics} & \cellcolor{ivtstride}43.0 \\
\midrule
{\textbf{ProstaTD}} \\ \midrule
TRCoRSurg~\cite{li2026trcorsurg} & CVPR 2026 & online & official folds & unknown & 37.5 \\
\rowcolor{gray!15}Ours & -- & online & official folds & \texttt{ivtmetrics} & \cellcolor{ivtstride}41.2 \\
\rowcolor{gray!15}Ours & -- & offline & official folds & \texttt{ivtmetrics} & \cellcolor{ivtstride}44.6 \\
\bottomrule
\end{tabular}
\caption{Published \apivt{} (\%) of surgical triplet recognition methods, as reported by their authors, with their venue, split and AP computation. Ours denotes our vision encoder pipeline with STRIDE.}
\label{tab:supp_published}
\end{table*}

\section{Published results}
\label{sec:supp-published}

\Cref{tab:supp_published} reports the published results on the three datasets, together with their split and AP computation. Not all of these results are directly comparable. CholecT50 is evaluated on its official cross-validation folds~\cite{nwoye2022splits} or on the split of Rendezvous~\cite{nwoye2022rendezvous}, and CholecT45 has its own folds. The computation of \apivt{} differs as well. We use \texttt{ivtmetrics} with all triplet classes, whereas some methods evaluate without the null triplets. We list the computation as unknown where we could not determine it. This explains why the results in \cref{tab:table1} can differ from the published ones. For example, CurConMix+ reaches 41.8 \apivt{} on CholecT45 with the evaluation of its public code, in line with the published 42.2, and 38.9 with our protocol.

We denote the performance of the foundation encoder pipeline~(\cref{sec:foundation}) as `Ours', with PL-Stitch for the online setting and TrackMAE for the offline setting~(\cref{tab:table2}).
\end{document}